%% file: main.tex
\documentclass[10pt,twocolumn,letterpaper]{article}

\usepackage{cvpr}
\usepackage[ruled,vlined,commentsnumbered,linesnumbered]{algorithm2e}
\usepackage{times}
\usepackage{epsfig}
\usepackage{graphicx}
\usepackage{amsmath}
\usepackage{amssymb}
\usepackage{epstopdf}
\usepackage{graphics}
\usepackage{color}
\usepackage{caption}
\usepackage{mathtools}
\usepackage{bbm}
\usepackage{lipsum}
\usepackage{subfig}
 \usepackage{array,multirow,graphicx}

\usepackage{epstopdf}

\usepackage[english]{babel}
\usepackage{booktabs} 
\usepackage{relsize}

\usepackage{pdfpages}

\usepackage[pagebackref=true,breaklinks=true,letterpaper=true,colorlinks,bookmarks=false]{hyperref}

\cvprfinalcopy 
 
\def\cvprPaperID{6129} 

\ifcvprfinal\fi
\begin{document}

\title{Learning a Neural Solver for Multiple Object Tracking}

\author{Guillem Bras{\'o}\thanks{Correspondence to: guillem.braso@tum.de.}
 \qquad \qquad Laura Leal-Taix{\'e} \vspace{0.3cm}
\\
Technical University of Munich\\
}

\maketitle

\thispagestyle{empty}
\global\csname @topnum\endcsname 0
\global\csname @botnum\endcsname 0

\begin{abstract}
Graphs offer a natural way to formulate Multiple Object Tracking (MOT) within the tracking-by-detection paradigm. 
However, they also introduce a major challenge for learning methods, as defining a model that can operate on such \textit{structured domain} is not trivial. As a consequence, most learning-based work has been devoted to learning better features for MOT, and then using these with well-established optimization frameworks.
In this work, we exploit the classical network flow formulation of MOT to define a fully differentiable framework based on Message Passing Networks (MPNs). By operating directly on the graph domain, our method can reason globally over an entire set of detections and predict final solutions. Hence, we show that learning in MOT does not need to be restricted to feature extraction, but it can also be applied to the data association step. 
We show a significant improvement in both MOTA and IDF1 on three publicly available benchmarks. Our code is available at \url{https://bit.ly/motsolv}.

\end{abstract}

\input{sections/introduction2.tex}

\begin{figure*}[h]
\vspace{-0.2cm}
        \centering
           \subfloat[Input]{%
              \includegraphics[height=5.8cm]{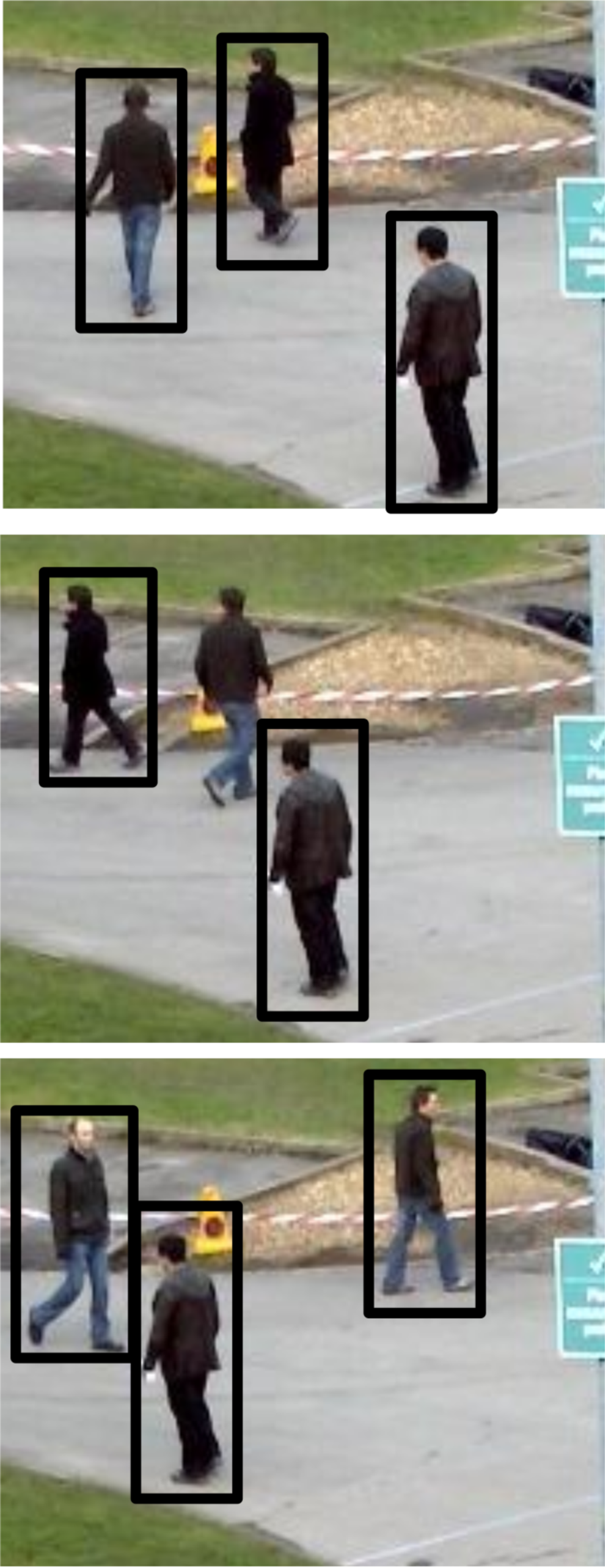}
              \label{input_fig}
           } 
           \hspace{0.3cm}
           \subfloat[Graph Construction + Feature Encoding]{%
              \includegraphics[height=5.8cm]{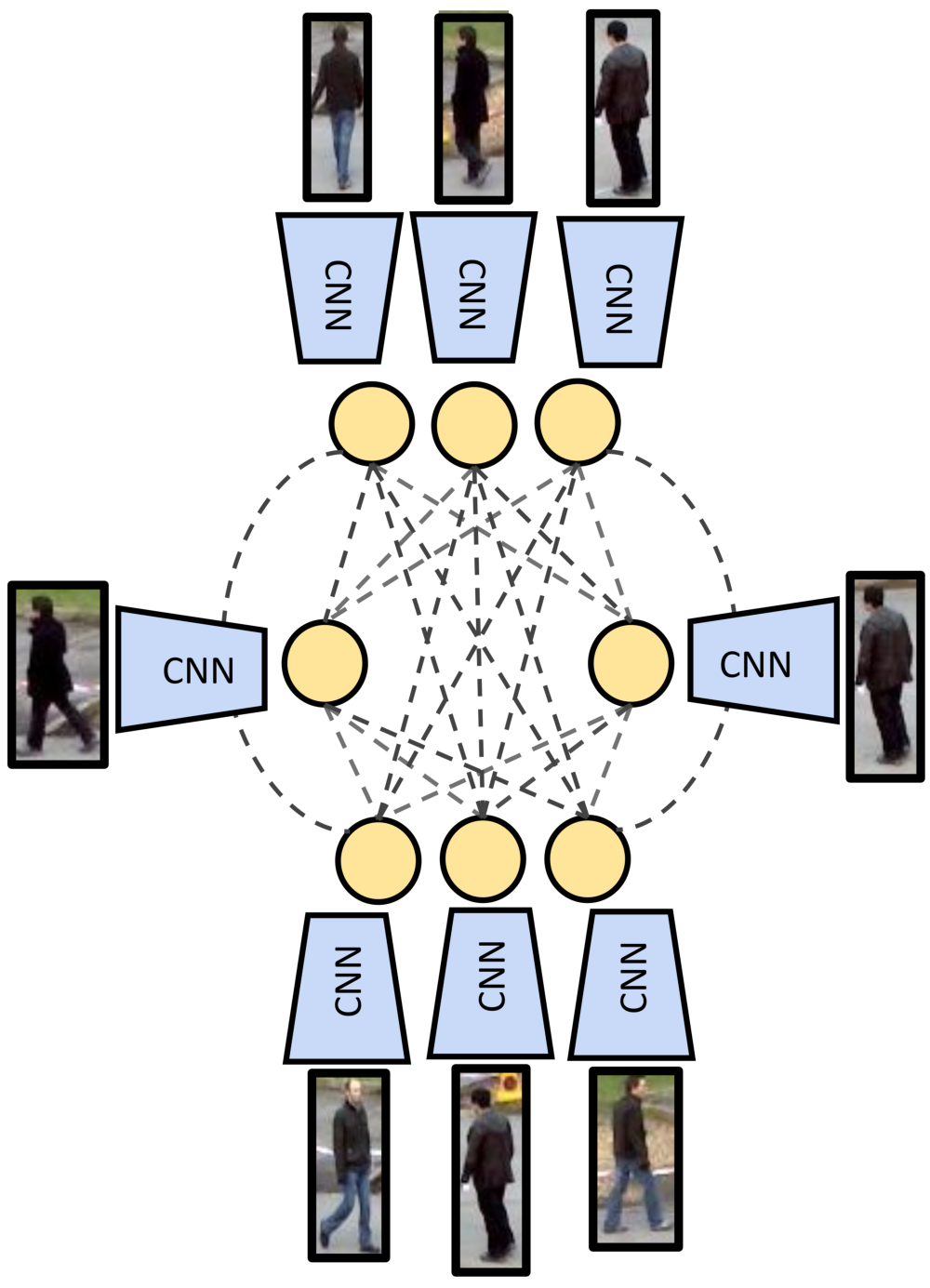}
              \label{graph_init}
           }
          \hspace{0.3cm}
           \subfloat[Neural Message Passing]{%
              \includegraphics[height=5.8cm]{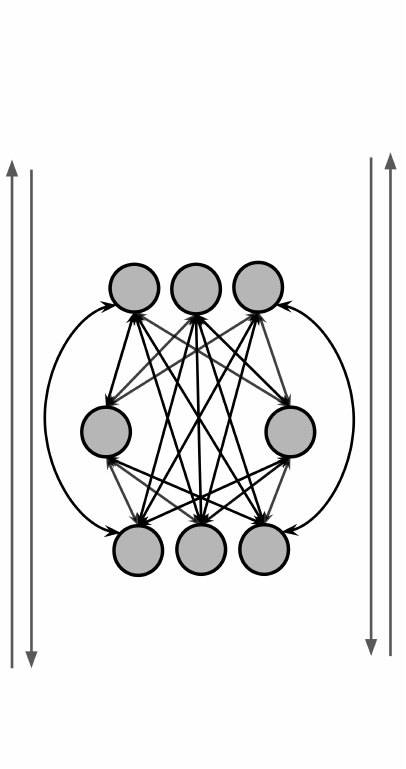} \label{mp_pipeline}
           }
          \hspace{0.3cm}
           \subfloat[Edge Classification]{%
              \includegraphics[height=5.8cm]{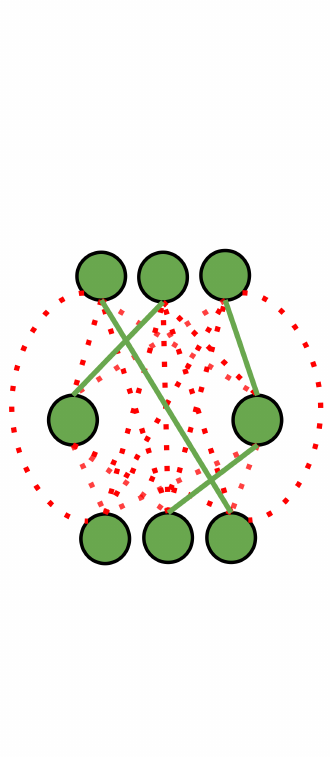} \label{edge_class_pipeline}
           }
          \hspace{0.3cm}
           \subfloat[Output]{%
              \includegraphics[height=5.8cm]{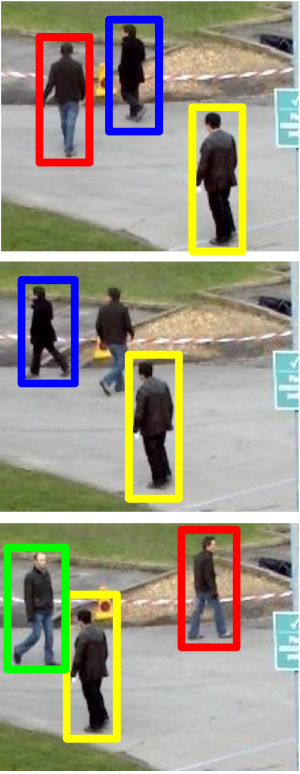} \label{output_pipeline}
           }           
\vspace{-0.2cm}
           \caption{Overview of our method. (a) We receive as input a set of frames and detections. (b) We construct a graph in which nodes represent detections, and all nodes at different frames are connected by an edge. (c) We initialize node embeddings in the graph with a CNN, and edge embeddings with an MLP encoding geometry information (not shown in figure). (c) The information contained in these embeddings is propagated across the graph for a fixed number of iterations through neural message passing. (d) Once this process terminates, the embeddings resulting from neural message passing are used to classify edges into active (colored with green) and non-active (colored with red). During training, we compute the cross-entropy loss of our predictions w.r.t. ground truth labels. (e) At inference, we follow a simple rounding scheme to binarize our classification scores and obtain final trajectories.}
           \label{pipeline}
\end{figure*}    

\input{sections/problem_formulation.tex}

\input{sections/tracking2.tex}
\input{sections/training.tex}

\input{sections/experiments.tex}

\section{Conclusion}

{We have demonstrated how to exploit the min-cost flow formulation of MOT to treat the entire tracking problem as a learning task. We have proposed a fully differentiable pipeline in which both feature extraction and data association can be jointly learned. At the core of our algorithm lies a message passing network with a novel time-aware update step that can capture the problem's graph structure. 
In our experiments, we have shown a clear performance improvement of our method with respect to previous state-of-the-art.
We expect our approach to open the door for future work to go beyond feature extraction for MOT, and focus, instead, on integrating learning into the overall data association task.}\\

\noindent\textbf{Acknowledgements.} This research was partially funded by the Humboldt Foundation through the Sofja Kovalevskaja Award.

\input{sections/bib.tex}

\clearpage
\appendix
\normalsize
\input{supplementary.tex}

\end{document}

%% file: sections/introduction2.tex

\section{Introduction}

Multiple object tracking (MOT) is the task of determining the trajectories of all object instances in a video. It is a fundamental problem in computer vision, with applications such as autonomous driving, biology, and surveillance. Despite its relevance, it remains a challenging task and a relatively unexplored territory in the context of deep learning. 

In recent years, \textit{tracking-by-detection} has been the dominant paradigm among state-of-the-art methods in MOT. This two step approach consists in first obtaining frame-by-frame object detections, and then linking them to form trajectories. 
While the first task can be addressed with learning-based detectors \cite{faster_rcnn, yolov2}, the latter, data association, is generally formulated as a graph partitioning problem \cite{tangcvpr2017,yucvpr2007,zhangcvpr2008,lealiccv2011,berclaztpami2011}. 
In this graph view of MOT, a node represents an object detection, and an edge represents the connection between two nodes. An active edge indicates the two detections belong to the same trajectory.
Solving the graph partitioning task, i.e., finding the set of active edges or trajectories, can also be decomposed into two stages. First, a cost is assigned to each edge in the graph encoding the likelihood of two detections belonging to the same trajectory. After that, these costs are used within a graph optimization framework to obtain the optimal graph partition.

Previous works on graph-based MOT broadly fall into two categories: those that focus on the graph formulation, and those that focus on learning better costs.  
In the first group, numerous research has been devoted to establishing complex graph optimization frameworks that combine several sources of information, with the goal of encoding high-order dependencies between detections \cite{People_Tracking, jCCpami2018, henscheltpami2016, JBNOT}. Such approaches often use costs that are \textit{handcrafted} to some extent. 
In the second group, several works adopt a simpler and easier to optimize graph structure, and focus instead on improving edge cost definition by leveraging deep learning techniques \cite{LealTaixeCVPR2014baseline, Son_2017_CVPR, Schulter_2017_CVPR, Zhu_2018_ECCV, sptn_iccv19}. 
By exploiting siamese convolutional neural networks (CNN), these approaches can encode reliable pairwise interactions among objects, but fail to account for high-order information in the scene. 
Overall, these two lines of work present a dilemma: should MOT methods focus on improving the graph optimization framework or the feature extraction?

We propose to combine both tasks into a unified learning-based solver that can: (i) learn features for MOT, and (ii) learn to provide a solution by reasoning over the entire graph.
To do so, we exploit the classical network flow formulation of MOT \cite{Zhang2008} to define our model. Instead of learning pairwise costs and then using these within an available solver, our method learns to directly predict final partitions of the graph into trajectories.  
Towards this end, we perform learning directly in the natural MOT domain, i.e., in the graph domain, with a message passing network (MPN). Our MPN learns to combine deep features into high-order information across the graph.  Hence, our method is able to account for global interactions among detections despite relying on a simple graph formulation.  
We show that our framework yields substantial improvements with respect to state of the art, without requiring heavily engineered features and being up to one order of magnitude faster than some traditional graph partitioning methods.

To summarize, we make the following {\bf contributions}:
\begin{itemize}
\item We propose a MOT solver based on message passing networks, which can exploit the natural graph structure of the problem to perform both feature learning as well as final solution prediction.
\item We propose a novel time-aware neural message passing update step inspired by classic graph formulations of MOT. 
\item We show significantly improved state-of-the-art results of our method in three public benchmarks.
\end{itemize}
 
 \section{Related Work}

Most state-of-the-art MOT works follow the tracking-by-detection paradigm which divides the problem into two steps: (i) detecting pedestrian locations independently in each frame, for which neural networks are currently the state-of-the-art~\cite{rennips2015,yolov2,sdpdetector}, and (ii) linking corresponding detections across time to form trajectories.

\noindent{\bf Tracking as a Graph Problem.} Data association can be done on a frame-by-frame basis for online applications \cite{breitensteiniccv2009, esscvpr2008, pellegriniiccv2009} or track-by-track \cite{berclazcvpr2006}. For video analysis tasks that can be done offline, batch methods are preferred since they are more robust to occlusions.
The standard way to model data association is by using a graph, where each
detection is a node, and edges indicates possible link among them. The data
association can then be formulated as maximum flow
\cite{berclaztpami2011} or, equivalently, minimum cost problem with either fixed costs based on distance
\cite{jiangcvpr2007,pirsiavashcvpr2011,zhangcvpr2008}, including motion models \cite{lealiccv2011}, or learned costs \cite{lealcvpr2014}. Both formulations can be solved optimally and efficiently. 
Alternative formulations typically lead to more involved optimization
problems, including minimum cliques~\cite{zamireccv2012},
general-purpose solvers, e.g., multi-cuts~\cite{tangcvpr2017}. 
A recent trend is to design ever more complex models which include other vision input such as reconstruction for multi-camera sequences \cite{lealcvpr2012,wucvpr2011}, activity recognition \cite{choieccv2012}, segmentation \cite{milancvpr2015}, keypoint trajectories \cite{choiiccv2015} or joint detection \cite{tangcvpr2017}.

\noindent{\bf Learning in Tracking.}
It is no secret that neural networks are now dominating the state-of-the-art in many vision tasks since \cite{krizhevskyImageNet} showed their potential for image classification. The trend has also arrived in the tracking community, where learning has been used primarily to learn a mapping from image to optimal costs for the aforementioned graph algorithms. The authors of \cite{lealcvprw2016} use a siamese network to directly learn the costs between a pair of detections, while a mixture of CNNs and recurrent neural networks (RNN) is used for the same purpose in \cite{Sadeghian_2017_ICCV}. More evolved quadruplet networks \cite{Son_2017_CVPR} or attention networks \cite{Zhu_2018_ECCV} have lead to improved results. In \cite{ristanicvpr2018}, authors showed the importance of learned reID features for multi-object tracking. 
All aforementioned methods learn the costs independently from the optimization method that actually computes the final trajectories. In contrast, \cite{kimaccv12,Wang2015b,Schulter_2017_CVPR} incorporate the optimization solvers into learning. The main idea behind these methods is that costs also need to be optimized for the solver in which they will be used. \cite{kimaccv12,Wang2015b, end_to_end_urtasun} rely on structured learning losses while \cite{Schulter_2017_CVPR} proposes a more general bi-level optimization framework. These works can be seen as similar to ours in spirit, given our common goal of incorporating the full inference model into learning for MOT. However, we follow a different approach towards this end: we propose to directly learn a solver and treat data association as a classification task, while their goal is to adapt their methods to perform well with non-learnable solvers. Moreover, all these works are limited to learning either pairwise costs \cite{end_to_end_urtasun, Schulter_2017_CVPR} or additional quadratic terms \cite{Wang2015b, kimaccv12} but cannot incorporate higher-order information as our method. Instead, we propose to leverage the common graph formulation of MOT as a domain in which to perform learning.

\noindent{\bf Deep Learning on Graphs.} 
Graph Neural Networks (GNNs) were first introduced in \cite{scarselli_graph_nn} as a generalization of neural networks that can operate on graph-structured domains. Since then, several works have focused on further developing and extending them by developing convolutional variants  \cite{Bruna2013SpectralNA, deferrand_cnns, kipf2016semi}. More recently, most methods were encompassed within a more general framework termed {neural message passing} \cite{Gilmer2017NeuralMP} and further extended in \cite{battaglia_graph_networks} as {graph networks}. Given a graph with some initial features for nodes and optionally edges, the main idea behind these models is to embed nodes (and edges) into representations that take into account not only the node's own features but also those of its neighbors in the graph, as well as the graph overall topology. These methods have shown remarkable performance at a wide variety of areas, ranging from chemistry \cite{Gilmer2017NeuralMP} to combinatorial optimization \cite{combinatorial_cnns}. Within vision, they have been successfully applied to problems such as human action recognition \cite{graph_nets_action_recognition}, visual question answering \cite{graph_nets_vqa} or single object tracking \cite{Gao_2019_CVPR}.

%% file: sections/problem_formulation.tex
\section{Tracking as a Graph Problem} \label{problem_formulation}

Our method's formulation is based on the classical min-cost flow view of MOT \cite{Zhang2008}. In order to provide some background and formally introduce our approach, we start by providing an overview of the network flow MOT formulation. We then explain how to leverage this framework to reformulate the data association task as a learning problem.

\subsection{Problem Statement}
In tracking-by-detection, we are given as input a set of object detections $\mathcal{O}=\{o_1, \dots , o_n\}$, where $n$ is the total number of objects for all frames of a video. Each detection is represented by $o_i = (a_i, p_i,t_i)$, where $a_i$ denotes the raw pixels of the bounding box, $p_i$ contains its 2D image coordinates and $t_i$ its timestamp. A trajectory is defined as a set of time-ordered object detections $T_i=\{{ o_{i_1}}, \dots , { o_{i_{n_i}}}\}$, where $n_i$ is the number of detections that form trajectory $i$. The goal of MOT is to find the set of trajectories $\mathcal{T}_*=\{T_1, \dots, T_m\}$, that best explains the observations $\mathcal{O}$. 

The problem can be modelled with an undirected graph $G=(V, E)$, where $V:=\{1, \dots ,n\}$, $E \subset V \times V $, and each node $i \in V$ represents a unique detection $o_i \in \mathcal{O}$. The set of edges $E$ is constructed so that every pair of detections, i.e., nodes, in different frames is connected, hence allowing to recover trajectories with missed detections. Now, the task of dividing the set of original detections into trajectories can be viewed as grouping nodes in this graph into disconnected components. Thus, each trajectory $T_i=\{{ o_{i_1}}, \dots , { o_{i_{n_i}}}\}$ in the scene can be mapped into a group of nodes  $\{i_1, \dots , i_{n_i}\}$ in the graph and vice-versa. 

\subsection{Network Flow Formulation} 
\label{sec:netflow}

In order to represent graph partitions, we introduce a binary variable for each edge in the graph. In the classical minimum cost flow formulation\footnote{We present a simplified version of the minimum cost flow MOT formulation~\cite{Zhang2008}. Specifically, we omit both sink and source nodes (and hence their corresponding edges) and we assume detection edges to be constant and 1-valued. We provide further details on our simplification and its relationship to the original problem in Appendix \ref{net_flow_formulation}.}  ~\cite{Zhang2008}, this label is defined to be 1 between edges connecting nodes that (i) belong to the same trajectory, and (ii) are temporally consecutive inside a trajectory; and 0 for all remaining edges. 

A trajectory $T_i=\{{ o_{i_1}}, \dots , { o_{i_{n_i}}}\}$ is equivalently denoted by the set of edges $\{(i_1, i_2), \dots ,(i_{n_i - 1}, i_{n_i})\} \subset E$, corresponding to its time-ordered path in the graph. We will use this observation to formally define the edge labels. For every pair of nodes in different timestamps, $(i, j) \in E$, we define a binary variable $y_{(i, j)}$ as:
$$
     y_{(i, j)} \coloneqq \begin{cases}
               1               & \exists T_k \in \mathcal{T}_* \text{ s.t. } (i, j) \in T_k \\
               0               & \text{otherwise.}
           \end{cases}
$$
An edge $(i,j)$ is said to be \textit{active} whenever $y_{(i, j)}=1$. 
We assume trajectories in $\mathcal{T}$ to be node-disjoint, i.e., a node cannot belong to more than one trajectory. Therefore, $\hat{y}$ must satisfy a set of linear constraints. For each node $i \in V$: 
\begin{align}
 \sum_{(j, i)\in E \text{ s.t. } t_i>t_j} y_{(j, i)} &\leq 1  \label{flow_in_constr} \\
\sum_{(i, k)\in E \text{ s.t. } t_i<t_k} y_{(i, k)} &\leq 1\label{flow_out_constr}
\end{align}

These inequalities are a simplified version of the \textit{flow conservation constraints}~\cite{Ahuja1993}. In our setting, they enforce that every node gets linked via an active edge to, at most, one node in past frames and one node in upcoming frames.

\subsection{From Learning Costs to Predicting Solutions}
In order to obtain a graph partition with the framework we have described, the standard approach is to first associate a cost $c_{(i, j)}$ to each binary variable $y_{(i, j)}$. This cost encodes the likelihood of the edge being active~\cite{Leal-Taixe:2014:CVPR, lealcvprw2016, Schulter_2017_CVPR}. The final partition is found by optimizing:
\begin{equation*}
\begin{array}{ll@{}ll}
\min_{y}  & \mathlarger{ \sum }_{(i, j) \in E} {c_{(i, j)}y_{(i, j)}}     &\\
\text{Subject to:} & \text{Equation } (\ref{flow_in_constr}) &\\
                &  \text{Equation } (\ref{flow_out_constr})      &\\                                               
&y_{(i,j)} \in \{0,1\}, & \quad (i, j) \in E
\end{array}
\end{equation*}
which can be solved with available solvers in polynomial time~\cite{Berclaz:2006:CVPR, networkflows}.

We propose to, instead, directly learn to predict which edges in the graph will be active, i.e., predict the final value of the binary variable $y$. 
To do so, we treat the task as a classification problem over edges, where our labels are the binary variables $y$. Overall, we exploit the classical network flow formulation we have just presented to treat the MOT problem as a fully learnable task.

%% file: sections/tracking2.tex
\section{Learning to Track with Message Passing Networks}
Our main contribution is a differentiable framework to train multi-object trackers as edge classifiers, based on the graph formulation we described in the previous section. 
Given a set of input detections, our model is trained to predict the values of the  binary \textit{flow} variables $y$ for every edge in the graph. 
Our method is based on a novel message passing network (MPN) able to capture the graph structure of the MOT problem. Within our proposed MPN framework, appearance and geometry cues are propagated across the entire set of detections, allowing our model to reason globally about the entire graph.

Our pipeline is composed of four main stages:

\noindent{\bf 1. Graph Construction:} Given a set of object detections in a video, we construct a graph where nodes correspond to detections and edges correspond to connections between nodes (Section \ref{sec:netflow}).

\noindent{\bf 2. Feature Encoding:} We initialize the node {\it appearance} feature embeddings from a convolutional neural network (CNN) applied on the bounding box image. 
For each edge, i.e., for every pair of detections in different frames, we compute a vector with features encoding their bounding box relative size, position and time distance. We then feed it to a multi-layer perceptron (MLP) that returns a {\it geometry} embedding (Section \ref{feature_encoding}).

\noindent{\bf 3. Neural Message Passing:} We perform a series of message passing steps over the graph. Intuitively, for each round of message passing, nodes share appearance information with their connecting edges, and edges share geometric information with their incident nodes. This yields updated embeddings for node and edges containing {\it higher-order} information that depends on the overall graph structure (Section \ref{vanilla_mpns} and \ref{time_aware_message_passing}).

\noindent{\bf 4. Training:} We use the final edge embeddings to perform binary classification into active/non-active edges, and train our model using the cross-entropy loss (Section \ref{inference}).\\

At test time, we use our model's prediction per edge as a continuous approximation (between 0 and 1) of the target \textit{flow} variables. We then follow a simple scheme to round them, and obtain the final trajectories.

For a visual overview of our pipeline, see Figure \ref{pipeline}.

\begin{figure*}[h]
        \centering
           \subfloat[Initial Setting]{%
              \includegraphics[height=6cm]{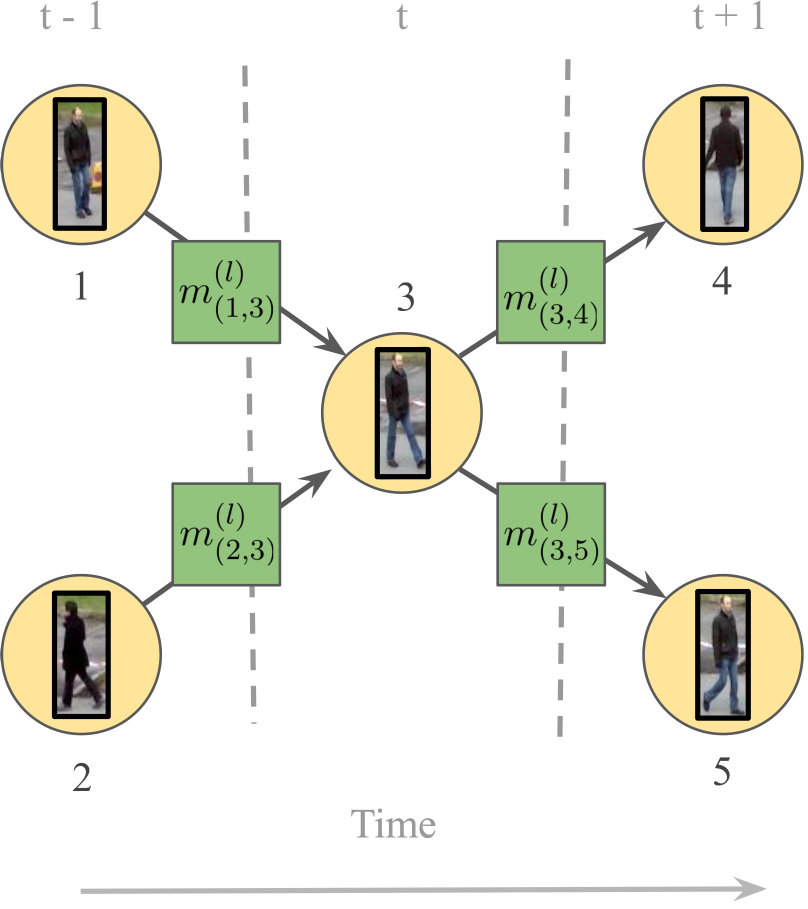}
              \label{setting}
           } 
           \hspace{0.3cm}
           \subfloat[Vanilla node update]{%
              \includegraphics[height=6cm]{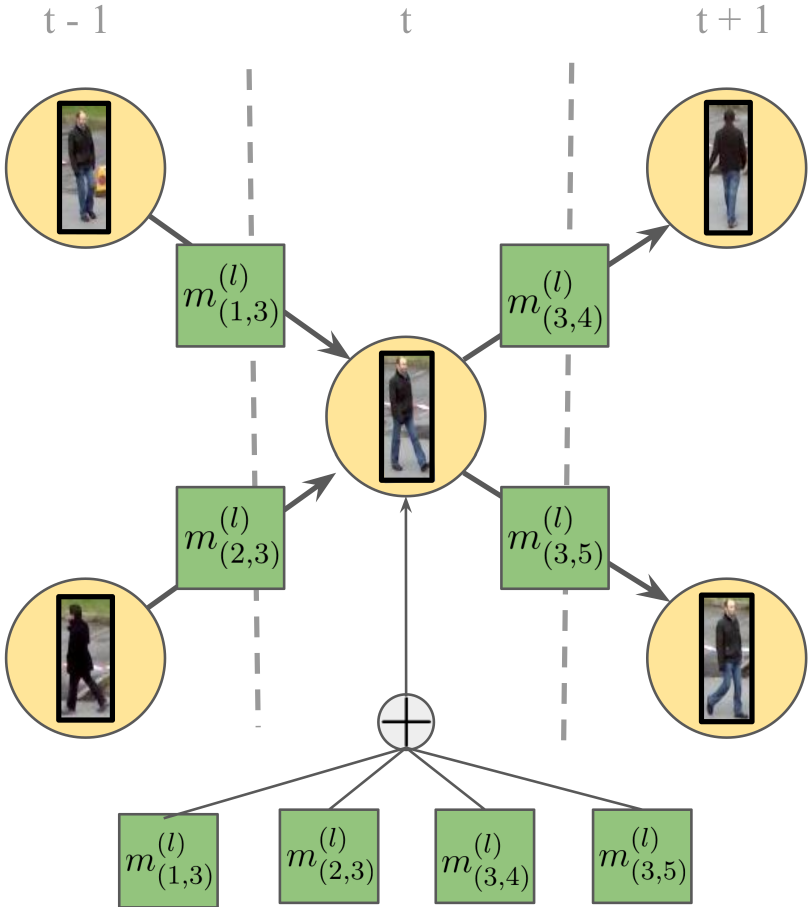}
              \label{vanilla_update}
           }
          \hspace{0.3cm}
           \subfloat[Time-aware node update]{%
              \includegraphics[height=6cm]{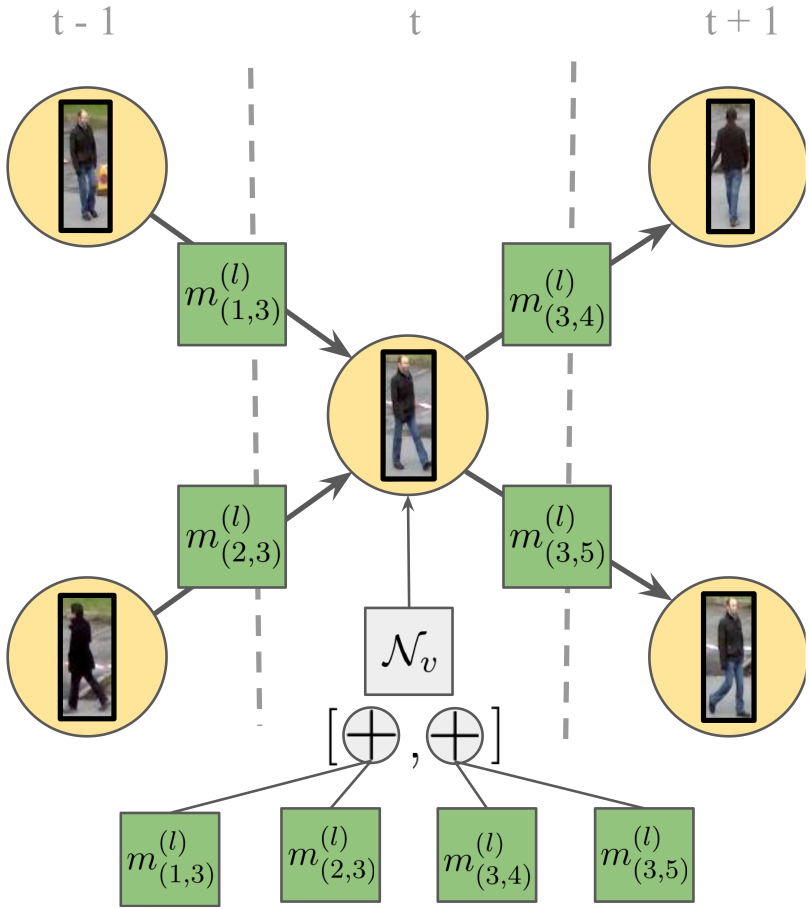} \label{time_aware_update}
           }
           \caption{Visualization of node updates during message passing. Arrow directions in edges show time direction. Note the time division in $t-1$, $t$, and $t+1$. In this case, we have $N_3^{past}=\{1, 2\}$ and $N_3^{fut}=\{4, 5 \}$. \ref{setting} shows the starting point after an edge update has been performed (equation \ref{node2edge}), and the intermediate node update embeddings (equation \ref{edge2node1}) have been computed. \ref{vanilla_update} shows the standard node update in vanilla MPNs, in which all neighbors' embeddings are aggregated jointly. \ref{time_aware_update} shows our proposed update, in which embeddings from past and future frames are aggregated separately, then concatenated and fed into an MLP to obtain the new node embedding.}\label{node_updates}
           \label{fig:default}
\end{figure*}    

\subsection{Message Passing Networks} \label{vanilla_mpns}

In this section, we provide a brief introduction to MPNs based on the work presented in \cite{Gilmer2017NeuralMP, kipf_icml2018, interaction_nets_battaglia, battaglia_graph_networks}. Let $G = (V, E)$ be a graph. Let $h_i^{(0)}$ be a node embedding for every $i\in V$, and $h_{(i, j)}^{(0)}$ an edge embedding for every $(i, j)\in E$. The goal of MPNs is to {learn} a function to propagate the information contained in nodes and edge feature vectors across $G$.

The propagation procedure is organized in {embedding updates} for edges and nodes, which are known as {\it message passing steps}~\cite{Gilmer2017NeuralMP}. In \cite{battaglia_graph_networks, kipf_icml2018, interaction_nets_battaglia}, each message passing step is divided, in turn, into two updates: one from from nodes to edges $(v\rightarrow e)$, and one from edges to nodes $(e\rightarrow v)$. 
The updates are performed sequentially for a fixed number of iterations $L$. For each $l\in \{1, \dots, L\}$, the general form of the updates is the following \cite{battaglia_graph_networks}:
\begin{align}
&(v\rightarrow e) &h_{(i, j)}^{(l)} &= \mathcal{N}_e\left([h_i^{(l-1)}, h_j^{(l-1)}, h_{(i, j)}^{(l-1)}]\right)  \label{node2edge}\\
&(e\rightarrow v) 
            &m_{(i, j)}^{(l)} &=  \mathcal{N}_v\left([h_i^{(l-1)}, h_{(i, j)}^{(l)} ]\right)  \label{edge2node1} \\
&            &h_i^{(l)} &= \Phi \left(\left \{ m_{(i, j)}^{(l)}   \right\}_{ j \in N_i}\right)  \label{edge2node2} 
\end{align}
Where $\mathcal{N}_e$ and $\mathcal{N}_v$ represent learnable functions, e.g., MLPs, that are shared across the entire graph. $[.]$ denotes concatenation, $N_i\subset V$ is the set of adjacent nodes to $i$, and $\Phi$ denotes an order-invariant operation, e.g., a summation, maximum or an average. 
Note, after $L$ iterations, each node contains information of all other nodes at distance $L$ in the graph. Hence, $L$ plays an analogous role to the {receptive field} of CNNs, allowing embeddings to capture context information.

\subsection{Time-Aware Message Passing} \label{time_aware_message_passing}
The previous message passing framework was designed to work on arbitrary graphs. 
However, MOT graphs have a very specific structure that we propose to exploit. Our goal is to encode a MOT-specific inductive bias in our network, specifically, in the node update step.

Recall the node update depicted in Equations~\ref{edge2node1} and \ref{edge2node2}, which allows each node to be compared with its neighbors and aggregate information from all of them to update its embedding with further context. 
Recall also the structure of our flow conservation constraints (Equations~\ref{flow_in_constr} and \ref{flow_out_constr}), which imply that each node can be connected to, at most, one node in future frames and another one in past frames. Arguably, aggregating all neighboring embeddings at once makes it difficult for the updated node embedding to capture whether these constraints are being violated or not (see Section \ref{ablation_study} for constraint satisfaction analysis). 

More generally, explicitly \textit{encoding} the temporal structure of MOT graphs into our MPN formulation can be a useful prior for our learning task. 
Towards this goal, we modify Equations~\ref{edge2node1} and \ref{edge2node2} into time-aware update rules by dissecting the aggregation into two parts: one over nodes in the past, and another over nodes in the future. 
Formally, let us denote the neighboring nodes of $i$ in future and past frames by ${N}^{fut}_i$ and ${N}^{past}_i$, respectively. 
Let us also define two different MLPs, namely, $\mathcal{N}^{fut}_v$ and $\mathcal{N}^{past}_v$. 
At each message passing step $l$ and for every node $i \in V$, we start by computing \textit{past} and \textit{future} edge-to-node embeddings for all of its neighbors $j\in {N}_i$ as:
\begin{align}
    m_{(i, j)}^{(l)} &= \begin{cases} \mathcal{N}^{past}_v\left([h_i^{(l-1)}, h_{(i, j)}^{(l)}, h_{(i)}^{(0)}]\right)  \text{ if } &j\in N^{past}_i \\
                  \mathcal{N}^{fut}_v \text{ }\left([h_i^{(l-1)}, h_{(i, j)}^{(l)}, h_{(i)}^{(0)}]\right) \text{   if } &j\in N^{fut}_i 
           \end{cases}
\end{align}
Note, the initial embeddings $h_{(i)}^{(0)}$ have been added to the computation\footnote{This skip connection ensures that our model does not \textit{forget} its initial features during message passing, and we apply it analogously with initial edge features in Equation~\ref{node2edge}.}. After that, we aggregate these embeddings separately, depending on whether they were in future or past positions with respect to $i$:
\begin{align}
    h_{i, past}^{(l)} &= \sum_{ j \in N^{past}_i} m^{(l)}_{(i, j)} \\
    h_{i, fut}^{(l)} &= \sum_{ j \in N^{fut}_i} m^{(l)}_{(i, j)} 
\end{align}

Now, these operations yield \textit{past} and \textit{future} embeddings $h_{i, past}^{(l)}$ and $h_{i, fut}^{(l)}$, respectively. We compute the final updated node embedding by concatenating them and feeding the result to one last MLP, denoted as $\mathcal{N}_v$:
\begin{align}
    h_i^{(l)} &=  \mathcal{N}_v([h_{i, past}^{(l)} , h_{i, fut}^{(l)} ])
\end{align}
We summarize  our time-aware update in Figure~\ref{time_aware_update}. As we demonstrate experimentally (see Section \ref{ablation_study}), this simple architectural design results in a significant performance improvement with respect to the \textit{vanilla} node update of MPNs, shown in Figure~\ref{vanilla_update}.  

%% file: sections/training.tex
\subsection{Feature Encoding} \label{feature_encoding}

The initial embeddings that our MPN receives as input are produced by other backpropagatable networks.

\noindent{\bf Appearance Embedding.} We rely on a convolutional neural network (CNN), denoted as  $\mathcal{N}_v^{enc}$, to \textit{learn} to extract a feature embeddings directly from RGB data. For every detection $o_i \in \mathcal{O}$, and its corresponding image patch $a_i$, we obtain $o_i$'s corresponding node embedding by computing $h_i^{(0)}:=\mathcal{N}_v^{enc}(a_i)$. 

\noindent{\bf Geometry Embedding.} We seek to obtain a representation that encodes, for each pair of detections in different frames, their relative position size, as well as distance in time. For every pair of detections $o_i$ and $o_j$ with timestamps $t_i \neq t_j$, we consider their bounding box coordinates parameterized by top left corner image coordinates, height and width, i.e., $(x_i, y_i, h_i, w_i)$ and $(x_j, y_j, h_j, w_j)$. We compute their relative distance and size as:
$$\left( \frac{2(x_j - x_i)}{h_i + h_j}, \frac{2(y_j - y_i)}{h_i + h_j}, \log\frac{h_i}{h_j}, \log\frac{w_i}{w_j}  \right)$$

We then concatenate this coordinate-based feature vector with the time difference  $t_j - t_i$ and relative appearance $\lVert\mathcal{N}_v^{enc}(a_j) - \mathcal{N}_v^{enc}(a_i) \rVert_2$ and feed it to a neural network $\mathcal{N}_e^{enc}$ in order to obtain the initial edge embedding $h_{(i, j)}^{(0)}$.

\subsection{Training and Inference} \label{inference}

\noindent{\bf Training Loss}.
To classify edges, we use an MLP with a sigmoid-valued single output unit, that we denote as $\mathcal{N}_e^{class}$. 
For every edge $(i, j)\in E$, we compute our prediction $\hat{y}_{(i,j)}^{(l)}$ by feeding the output embeddings of our MPN at a given message passing step $l$, namely $h_{(i, j)}^{(l)}$, to $\mathcal{N}_e^{class}$. 
For training, we use the binary cross-entropy of our predictions over the embeddings produced in the last message passing steps, with respect to the target flow variables $y$:

{\footnotesize
\begin{align}
\mathcal{L} = \frac{-1}{|E|}\sum_{l = l_0}^{l = L}\sum_{(i,j) \in E} w \cdot y_{(i, j)} \log(\hat{y}_{(i,j)}^{(l)}) +
(1 - y_{(i, j)}) \log(1 - \hat{y}_{(i,j)}^{(l)})
\end{align}}%

\noindent where $l_0\in \{1, \dots, L\}$ is the first message passing step at which predictions are computed, and $w$ denotes a positive scalar used to weight 1-valued labels to account for the high imbalance between active and inactive edges. 

\noindent{\bf Inference.}
During inference, we interpret the set of output values obtained from our model at the last message passing step as the solution to our MOT problem, i.e., the final value for the indicator variables $y$. 
Since these predictions are the output of a sigmoid unit, their values are between 0 and 1. 
An easy way to obtain hard $0$ or $1$ decisions is to binarize the output by thresholding. 
However, this procedure does not generally guarantee that the flow conservation constraints in Equations~\ref{flow_in_constr} and \ref{flow_out_constr} are preserved.
In practice, thanks to the proposed time-aware update step, our method will satisfy over $98\%$ of the constraints on average when thresholding at 0.5. After that, a simple greedy rounding scheme suffices to obtain a feasible binary output. The exact optimal rounding solution can also be obtained efficiently with a simple linear program. We explain both procedures in Appendix \ref{rounding_section}.

%% file: sections/experiments.tex
\section{Experiments} \label{experiments_section}

In this section, we first present an ablation study to better understand the behavior of our model. We then compare to published methods on three datasets, and show state-of-the-art results. All experiments are done on the MOTChallenge pedestrian benchmark.

\noindent{\bf Datasets and Evaluation Metrics.}
The multiple object tracking benchmark MOTChallenge~\footnote{The official MOTChallenge web page is available at ~\url{https://motchallenge.net}.} consists of several challenging pedestrian tracking sequences, with frequent occlusions and crowded scenes. 
The challenge includes three separate tracking benchmarks, namely {\it 2D MOT 2015}~\cite{lealarxiv2015}, {\it MOT16}~\cite{milanarxiv2016} and {\it MOT17}~\cite{milanarxiv2016}. They contain sequences with varying viewing angle, size and number of objects, camera motion and frame rate.
For all challenges, we use the detections provided by MOTChallenge to ensure a fair comparison with other methods. 
The benchmark provides several evaluation metrics. The Multiple Object Tracking Accuracy (MOTA)~\cite{clear} and ID F1 Score (IDF1)~\cite{ristanieccvw2016} are the most important ones, as they quantify two of the main aspects of multiple object tracking, namely, object coverage and identity preservation.

\subsection{Implementation Details}

\noindent{\bf Network Models.} For the network $\mathcal{N}_v^{enc}$ used to encode detections appearances (see section  \ref{feature_encoding}), we employ a ResNet50\cite{He2016DeepRL} architecture pretrained on ImageNet \cite{imagenet_cvpr09}, followed by global average pooling and two fully-connected layers to obtain embeddings of dimension 256.

We train the network for the task of ReIdentification (ReID) jointly on three publicly available datasets: Market1501\cite{market_dataset}, CUHK03\cite{cuhk03_dataset} and DukeMTMC\cite{ristanieccvw2016}. Note that using external ReID datasets is a common practice among MOT methods \cite{People_Tracking,Kim_2018_ECCV, maACCV2019}.
Once trained, three new fully connected layers are added after the convolutional layers to reduce the embedding size of $\mathcal{N}_v^{enc}$ to 32.
The rest of the encoder and classifier networks are MLPs and their exact architectures are detailed in Table \ref{tab:layer_sizes} in the supplementary material. 

\noindent{\bf Data Augmentation.} To train our network, we sample batches of 8 graphs. Each graph corresponds to small clips of 15 frames sampled at 6 frames per second for static sequences, and 9 frames per second for those with a moving camera. We do data augmentation by randomly removing nodes from the graph, hence simulating missed detections, and randomly shifting bounding boxes. 

 \noindent{\bf Training.} We have empirically observed that additional training of the ResNet blocks provides no significant increase in performance, but carries a significantly larger computational overhead. 
 Hence, during training, we freeze all convolutional layers and train jointly all of the remaining model components. 
 We train for 15000 iterations with a  learning rate $3 \cdot 10^{-4}$, weight decay term $10^{-4}$ and an Adam Optimizer with $\beta_1$ and $\beta_2$ set to $0.9$ and $0.999$, respectively. 

\noindent{\bf Batch Processing.} We process videos offline in batches of $15$ frames, with $14$ overlapping frames between batches to ensure that the maximum time distance between two connected nodes in the graph remains stable along the whole graph. 
We prune graphs by connecting two nodes only if both are among the top-$K$ mutual nearest neighbors (with $K = 50$) according to the ResNet features.  
Each batch is solved independently by our network, and for overlapping edges between batches, we average the predictions coming from the all graph solutions before the rounding step.
To fill gaps in our trajectories, we perform simple bilinear interpolation along missing frames.

\noindent{\bf Baseline.}
Recently, \cite{tracktor} has shown the potential of detectors for simple data association, establishing a new baseline for MOT. To exploit it, we preprocess all sequences by first running \cite{tracktor} on public detections, which allows us to be  fully comparable to all methods on MOTChallenge. One key drawback of \cite{tracktor} is its inability to fill in gaps, nor properly recover identities through occlusions. As we will show, this is exactly where our method excels. In Appendix \ref{graph_meth_comparison_section}, we show additional results without \cite{tracktor}.

\noindent{\bf Runtime.} We build our graph on the output of \cite{tracktor}. Hence, we take also its runtime into account. Our method, on its own, runs at 35fps, while \cite{tracktor} without the added re-ID head runs at 8fps, which gives the reported average of 6.5fps.

\subsection{Ablation Study} \label{ablation_study}
In this section, we aim to answer three main questions towards understanding our model. Firstly, we compare the performance of our time-aware neural message passing updates with respect to the time-agnostic vanilla node update described in \ref{vanilla_mpns}.
Secondly, we assess the impact of the number of message passing steps in network training to the overall tracking performance.
Thirdly, we investigate how different information sources, namely, appearance embeddings from our CNN and relative position information, affect different evaluation metrics. 
%

\noindent{\bf Experimental Setup}. We conduct all of our experiments with the training sequences of the MOT15 and MOT17 datasets. 
To evaluate our models, we split MOT17 sequences into three sets, and use these to test our models with 3-fold cross-validation. We then report the best overall MOT17 metrics obtained during validation. See Appendix \ref{cross_val_splits} for more details. In order to provide a fair comparison with configurations that show poor constraint satisfaction, we use exact rounding via a linear program in all experiments (see Section \ref{inference}).

\noindent{\bf Time-Aware Message Passing}. We investigate how our proposed time-aware node update affects performance. For a fair comparison, we perform hyperparameter search for our baseline. Still, we observe a significant improvement in almost all metrics, including close to 3 points in IDF1. 
As we expected, our model is particularly powerful at linking detections, since it exploits neighboring information and graph structure, making the decisions more robust, and hence producing significantly less identity switches. 
We also report the percentage of constraints that are satisfied when directly thresholding our model's output values at 0.5. Remarkably, our method with time-aware node updates is able to produce almost completely feasible results automatically, i.e., 98.8\% constraint satisfaction, while the baseline has only 82.1\% satisfaction. This demonstrates its ability to capture the MOT problem structure.

\input{tables/ablation_arch} \label{ablation_arch}

\begin{figure} 

\includegraphics[width=0.50\textwidth]{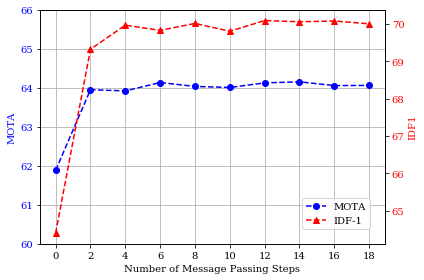}
\centering
\caption{We report the evolution of IDF-1 and MOTA when training networks with an increasing number of message passing steps.} \label{num_mp_steps}
\end{figure}

\noindent{\bf Number of Message Passing Steps}. 
Intuitively, increasing the number of message passing steps $L$ allows each node and edge embedding to encode further context, and gives edge predictions the ability to be iteratively refined.
Hence, one would expect higher values to yield better performing networks. We test this hypothesis in Figure \ref{num_mp_steps} by training networks with a fixed number of message passing steps, from 0 to 18. We use the case $L=0$ as a baseline in which we train a binary classifier on top of our initial edge embeddings, and hence, no contextual information is used.
As expected, we see a clear upward tendency for both IDF-1 and MOTA. Moreover, we observe a steep increase in both metrics from zero to two message passing steps, which demonstrates that the biggest improvement is obtained when switching from pairwise to high-order features in the graph. 
 We also note that the upwards tendency stagnates around six message passing steps, and shows no improvement after twelve message passing steps. Hence, we use $L=12$ in our final configuration.

\noindent{\bf Effect of the Features}. Our model receives two main streams of information: (i) appearance information from a CNN, and (ii) geometry features from an MLP encoding relative position between detections. 
We test their usefulness by experimenting with combinations of three groups of features for edges: time difference, relative position and the euclidean distance in CNN embeddings between the two bounding boxes. Results are summarized in Table \ref{tab:ablation_features_table}.
We highlight the fact that relative position seems to be a key component to overall performance since its addition yields the largest relative performance increase.
%
Nevertheless, CNN features are powerful to reduce the number of false positives and identity switches and hence, we use them in our final configuration. 

\input{tables/ablation_features} \label{ablation_features_table}

\subsection{Benchmark Evaluation}
We report the metrics obtained by our model on the MOT15, MOT16 and MOT17 datasets in Table~\ref{tab:mot}. 
Our method obtains state-of-the-art results by a large margin on all challenges, improving especially the IDF1 measure by 11, 6.4, and 6.6 percentage points, respectively, which demonstrates our method's strong performance in identity preservation. 
We attribute this performance increase to the ability of our message passing architecture to collect higher-order information. Taking into consideration neighbors' information when linking trajectories allows our method to make \textit{globally informed} predictions, which leads inevitably to less identity  switches. Moreover, we also achieve more trajectory coverage, represented by an increase in Mostly Tracked (MT) trajectories of up to 9 percentage points. 
It is worth noting the big performance improvement with respect to previous graph partitioning methods (shown as (G) in Table ~\ref{tab:mot}), which often use expensive optimization schemes. Not only do we surpass them by a large margin, but we are also up to one order of magnitude faster than some of them, e.g. \cite{jCCpami2018}.  In Appendix \ref{graph_meth_comparison_section}, we show a more detailed comparison. 

\input{tables/mot} \label{mot_metrics_results}

%% file: tables/ablation_arch.tex

\begin{table}
\center
\tabcolsep=0.11cm

    \resizebox{\columnwidth}{!}{
    \begin{tabular}{l c  c c c c c c c c}
     \toprule
     Arch. & MOTA $\uparrow$ & IDF1 $\uparrow$ & MT $\uparrow$ & ML $\downarrow$ & FP $\downarrow$ & FN $\downarrow$ & ID Sw. $\downarrow$ &Constr. $\uparrow$\\ [0.5ex] 
     \midrule
     Vanilla  & 63.0  & 67.3 & 586 &  372 & 41281 & 119542 & 1022 & 82.1 \\
     T. aware  & 64.0 & 70.0 & 648 &  362 & 6169 & 114509 & 602   & 98.8 \\

    \end{tabular}}

\caption{We investigate how our proposed update improves tracking performance with respect to a vanilla MPN. \textit{Vanilla} stands for a basic MPN, \textit{T. aware} denotes our proposed time-aware update. The metric \textit{Constr} refers to the percentage of flow conservation constraints satisfied on average over entire validation sequences.}
\vspace{-0.2cm}
\label{tab:mot}

\end{table}

%% file: tables/ablation_features.tex

\begin{table}
\center
\tabcolsep=0.11cm

    \resizebox{\columnwidth}{!}{
    \begin{tabular}{l l  c c c c c c c c}
     \toprule
     Edge Feats. & MOTA $\uparrow$ & IDF1 $\uparrow$ & MT $\uparrow$ & ML $\downarrow$ & FP $\downarrow$ & FN $\downarrow$ & ID Sw. $\downarrow$ \\ [0.5ex] 
     \midrule

     Time & 58.8 &  52.6 &  529 &  372 &  13127 &  122800 &  2962 \\

     Time+Pos & 63.6 &  68.7 &  631 &  365 &   6308 &  115506 &   895 \\
    
     Time+Pos+CNN  & 64.0 & 70.0 & 648 &  362 & 6169 & 114509 & 602    \\

     \midrule

    \end{tabular}}

\caption{We explore combinations of three sources of information for edge features: time difference in seconds (Time), relative position features (Pos) and the Euclidean distance between CNN embeddings of the two detections (CNN).}
\vspace{-0.2cm}
\label{tab:ablation_features_table}
\end{table}

%% file: tables/mot.tex

\begin{table}
\center
\tabcolsep=0.11cm

    \resizebox{\columnwidth}{!}{
    \begin{tabular}{l c c c c c c c c c }
     \toprule
     Method & MOTA $\uparrow$ & IDF1 $\uparrow$ & MT $\uparrow$ & ML $\downarrow$ & FP $\downarrow$ & FN $\downarrow$ & ID Sw. $\downarrow$  & Hz  $\uparrow$\\ [0.5ex] 
     \midrule
     \multicolumn{8}{c}{2D MOT 2015~\cite{lealarxiv2015}    } \\
     \midrule
     Ours & \textbf{51.5} &  \textbf{58.6} & \textbf{31.2} &  \textbf{25.9}  & 7620 & \textbf{21780} & \textbf{375} & 6.5 \\
     Tracktor  \cite{tracktor} & 46.6 &  47.6 & 18.2 &  27.9  & 4624 & 26896 & 1290 & 1.8\\
     KCF~\cite{wacv_soa}  (G) & 38.9 & 44.5 & 16.6 & 31.5 & 7321 & 29501 & 720 & 0.3 \\     
     AP\_HWDPL\_p~\cite{ChenASZB17}  (G) & 38.5 & 47.1 & 8.7 & 37.4 & \textbf{4005} & 33203 & 586  & 6.7\\
     STRN~\cite{spatio_temporalrelation_networks} & 38.1 & 46.6 & 11.5 & 33.4 & 5451 & 31571 & 1033  & 13.8\\
     AMIR15~\cite{SadeghianAS17} & 37.6 & 46.0 & 15.8 & 26.8 & 7933 & 29397 & 1026 & 1.9 \\
     JointMC~\cite{jCCpami2018} (G) & 35.6 & 45.1 & 23.2 & 39.3 & 10580 & 28508 & 457 & 0.6\\

      DeepFlow\cite{Schulter_2017_CVPR} (G) & 26.8 & -- & -- & -- & -- & -- & -- & --\\
     \midrule
     \multicolumn{8}{c}{MOT16~\cite{milanarxiv2016}} \\
     \midrule
     Ours &  \textbf{58.6} & \textbf{61.7} & \textbf{27.3} &  \textbf{34.0} & 4949 & \textbf{70252} &  \textbf{354} & 6.5 \\
     Tracktor \cite{tracktor} & 56.2 &  54.9  & 20.7 & 35.8 &  \textbf{2394} & 76844 & 617 & 1.8\\

     NOTA \cite{aggregate_Track_app} (G)  & 49.8 &	55.3 &	17.9 & 37.7 & 7428 & 83614 & 614 & --\\
     
     HCC~\cite{maACCV2019}  (G)& 49.3 & 50.7 & 17.8 & 39.9 & 5333 & 86795 & 391  & 0.8\\

     LMP~\cite{TangAAS17}  (G)& 48.8 & 51.3 & 18.2 & 40.1 & 6654 & 86245 & 481  & 0.5\\
     KCF~\cite{wacv_soa}  (G) & 48.8 & 47.2 & 15.8 & 38.1 & 5875 & 86567 & 906 & 0.1 \\
     
     GCRA~\cite{MaYYZZJX18} & 48.2 & 48.6 & 12.9 & 41.1 & 5104 & 88586 & 821 & 2.8 \\
     FWT~\cite{HenschelLCR17}  (G) & 47.8 & 44.3 & 19.1 & 38.2 & 8886 & 85487 & 852 & 0.6 \\

     \midrule
     \multicolumn{8}{c}{ MOT17~\cite{milanarxiv2016}   } \\
     \midrule
          Ours & \textbf{58.8} & \textbf{61.7} & \textbf{28.8} & \textbf{33.5} & 17413 & \textbf{213594} & \textbf{1185} & 6.5 \\
     Tracktor\cite{tracktor}     & 56.3 & 55.1 & 21.1 & 35.3 & \textbf{8866} & 235449 & 1987 & 1.8\\
     JBNOT \cite{JBNOT}  (G)& 52.6 & 50.8 & 19.7 & 35.8 & 31572 & 232659 & 3050 & 5.4 \\

     FAMNet \cite{famnet}    & 52.0 & 48.7 & 19.1 & 33.4 & 14138 & 253616 & 3072 & --\\     
     eHAF\cite{sheng2018}  (G)    & 51.8 & 54.7 & 23.4 & 37.9 & 33212 & 236772 & 1834 & 0.7\\
     NOTA \cite{aggregate_Track_app}  (G) & 51.3 &	54.7 &	17.1 & 35.4 & 20,148 & 252,531 & 2,285 & --\\
     FWT~\cite{HenschelLCR17}  (G)& 51.3 & 47.6 & 21.4 & 35.2 & 24101 & 247921 & 2648 & 0.2	 \\
     
     jCC~\cite{jCCpami2018}  (G) & 51.2 & 54.5   & 20.9 & 37.0 & 25937 & 247822 & 1802 &1.8\\
     \bottomrule
    \end{tabular}}

\caption{Comparison of our method with state-of-the art. We set new state-of-the art results by a significant margin in terms of MOTA and especially IDF1. Our learned solver is more accurate while being significantly faster than most graph partitioning methods, indicated with (G).}
\vspace{-0.2cm}
\label{tab:mot}

\end{table}

%% file: sections/bib.tex
{\small
\bibliographystyle{ieee}
\bibliography{pubs}
}

%% file: supplementary.tex

\noindent{\LARGE{\textbf{Supplementary Material}}}

\maketitle

\thispagestyle{empty}
\global\csname @topnum\endcsname 0
\global\csname @botnum\endcsname 0

\input{sections/net_flow_formulation.tex}

\input{sections/rounding.tex}

\input{tables/layer_sizes.tex} 
\input{sections/further_implementation_details.tex}

\input{sections/no_tracktor_metrics.tex}






%
%
%


%% file: sections/net_flow_formulation.tex
\section{Network Flow Formulation} \label{net_flow_formulation}
\begin{figure*}[h]
        \centering
           \subfloat[Tracking Output]{%
              \includegraphics[height=6cm]{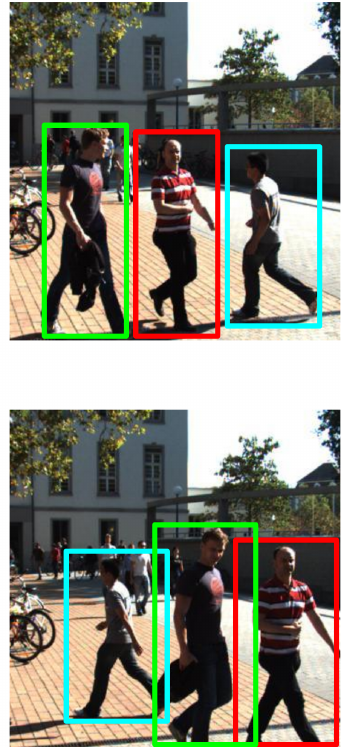}
              \label{tracking_out}
           } 
           \hspace{0.3cm}
           \subfloat[Classical Flow Formulation]{%
              \includegraphics[height=6cm]{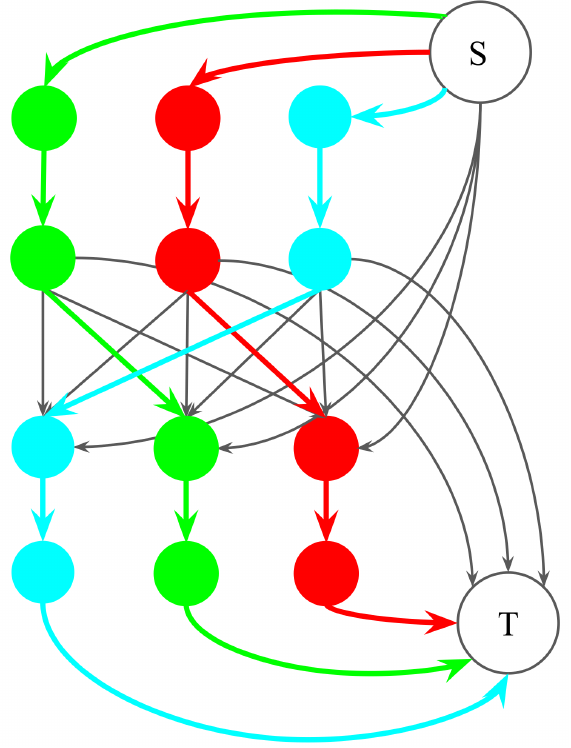}
              \label{classical_view}
           }
          \hspace{0.3cm}
           \subfloat[Simplified Flow Formulation]{%
              \includegraphics[height=6cm]{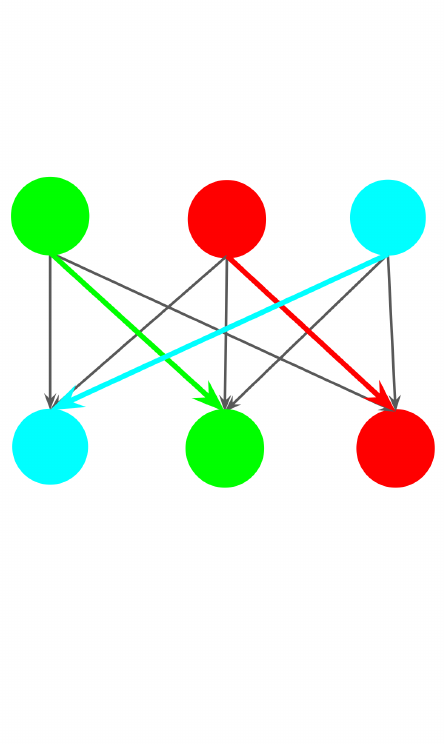} \label{simplified_view}
           }

           \caption{Overview of the simplification of the classical min-cost flow formulation of MOT used in our method. \ref{tracking_out} shows two frames with different trajectories indicated by different bounding box color. \ref{classical_view} shows the classical flow-based view of the scenario: active detections are displayed with \textit{detection} edges, and start (resp. ends) of trajectories are indicated with connections to the source (S) (resp. source (T)) node. \ref{simplified_view} shows the formulation used in our approach: no sink nor source nodes are used, and active detections are represented with a single node.} \label{simplified_flow_form}
           
\end{figure*}

In this section, we detail how the network flow-based formulation we present in the main paper is related to the original one proposed in \cite{Zhang2008}. See Figure \ref{simplified_flow_form} for an overview.

\subsection{Active and Inactive Detections}
Let $G = (V, E)$ be a graph representing a multiple object tracking (MOT) problem. In our method's graph formulation, we use nodes and edges to represent, respectively, detections and possible links forming trajectories among them. 
Moreover, we assign a binary variable $y_{(i, j)}$ to every edge $(i, j) \in E$ to represent whether the link between detections $i$ and $j$ is active or not. Nodes, i.e., detections, do not have any variable assigned to them, and we assume that all detections in the graph are correct. 
That is, we assume that there are no false positives in the graph\footnote{We can make this assumption because we can easily filter false positives from our graphs during pre-processing and post-processing (see Appendix \ref{futher_impl_details_section}).}.

In the general min-cost flow formulation, instead, detections are also represented with edges and they are assigned, in turn, another binary variable that indicates whether the detection is a true or false positive. 
Formally, for the $i$th input detection this binary variable is denoted as $y_i$, and its value is one if the detection is a true positive, i.e., it is active, and zero otherwise, i.e., it is inactive.\\

By allowing nodes to be inactive, the flow conservation constraints \ref{flow_in_constr} and \ref{flow_out_constr} we described in the main paper:
\begin{align}
\sum_{(j, i)\in E \text{ s.t. } t_i>t_j} y_{(j, i)} &\leq 1 \tag{\ref{flow_in_constr}} \\
\sum_{(i, k)\in E \text{ s.t. } t_i<t_k} y_{(i, k)} &\leq 1 \tag{\ref{flow_out_constr}}
\end{align}
are no longer sufficient. Instead, these constraints need to capture that, if a detection is inactive, both its incoming and outgoing flows need to be zero. This is achieved by replacing the right-hand-side of these inequalities with the binary variable $y_i$:
\begin{align}
 \sum_{(j, i)\in E \text{ s.t. } t_i>t_j} y_{(j, i)} &\leq y_i  \label{flow_in_constr_w_node} \\
\sum_{(i, k)\in E \text{ s.t. } t_i<t_k} y_{(i, k)} &\leq y_i\label{flow_out_constr_w_node}
\end{align}
Observe that whenever $y_i=0$, all edges entering and leaving detection $i$ need to be inactive. In contrast, when $y_i = 1$, i.e., the detection is active, these constraints are equivalent to the ones we use.

\subsection{Source and Sink Nodes}
In the classical min-cost flow formulation of MOT \cite{Zhang2008}, there are two special nodes: \textit{source} and \textit{sink}. Every detection $i$ is connected to both of them, and the resulting edge receives a binary variable denoted as $y_{en, i}$ and $y_{ext, i}$, respectively. These are used to indicate whether a trajectory starts or ends at $i$. Observe that $y_{en, i} = 1$ if, and only if, there is no detection $j$ in a past frame such that $y_{i, j} = 1$, and analogously for $y_{ext, i} = 1$. Hence, these variables can be used to transform inequalities \ref{flow_in_constr_w_node} and \ref{flow_out_constr_w_node} into equalities as:

\begin{align}
 y_{en, i} + \sum_{(j, i)\in E \text{ s.t. } t_i>t_j} y_{(j, i)} &= y_i  \label{full_flow_in_constr_w_node} \\
y_{ext, i} + \sum_{(i, k)\in E \text{ s.t. } t_i<t_k} y_{(i, k)} &= y_i\label{full_flow_out_constr_w_node}
\end{align}
which yield  the flow conservation constraints introduced in \cite{Zhang2008}.
In the original min-cost flow formulation, these edges are assigned a handcrafted cost indicating the \textit{price} of starting or ending a trajectory. If this cost is set to zero, one can think about $y_{en, i}$ and $y_{ext, i}$ as \textit{slack} variables.

\subsection{Overview of our Simplification}

To summarize, in our method we simplify the min-cost flow formulation by eliminating two elements of the classical one: detection edges and sink and source nodes. The first choice allows us to decouple the data association problem from the identification of incorrect detections. Since the latter task can be easily tackled in our pre-processing and post-processing routines (see Appendix \ref{futher_impl_details_section}), we can simplify our graph formulation and allow our network to focus on the task of edge classification. As for not using sink and source nodes, in our method there is no need for such \textit{special} variables. Instead, the start (resp. end) of a trajectory is naturally indicated by the absence of active incoming (resp. outgoing) edges to a node. Overall, we simplify the min-cost flow MOT formulation and reduce it to its most essential component: association edges. As a result, we obtain a setting that is suited for our message passing network to operate and effectively learn our task at hand.


%% file: sections/rounding.tex
\section{Rounding Solutions}\label{rounding_section}
\begin{figure*}[h]
        \centering
           \subfloat[Output Graph Binarized by Thresholding]{%
              \includegraphics[width=0.2\textwidth]{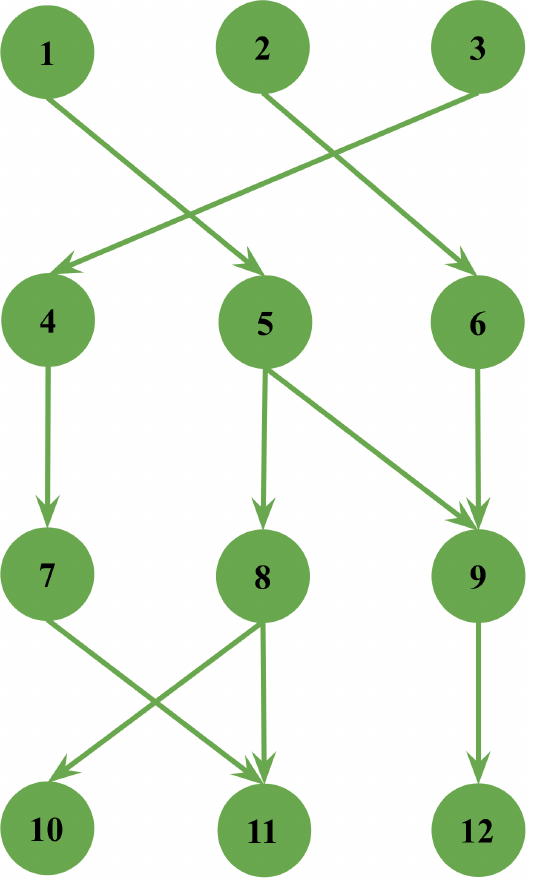}
              \label{trehsold_result}
           } 
          \hspace{0.5cm}
           \subfloat[Subgraph of Violated Constraints]{%
              \includegraphics[width=0.2\textwidth]{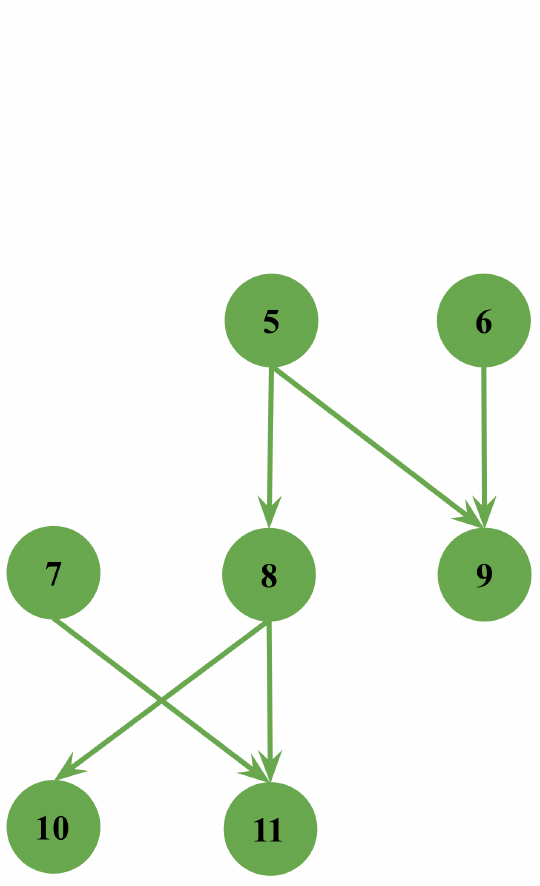}
              \label{viol_subgraph}
           }
          \hspace{0.5cm}
           \subfloat[Subgraph of Violated Constraints with Exact / Greedy Rounding]{%
              \includegraphics[width=0.2\textwidth]{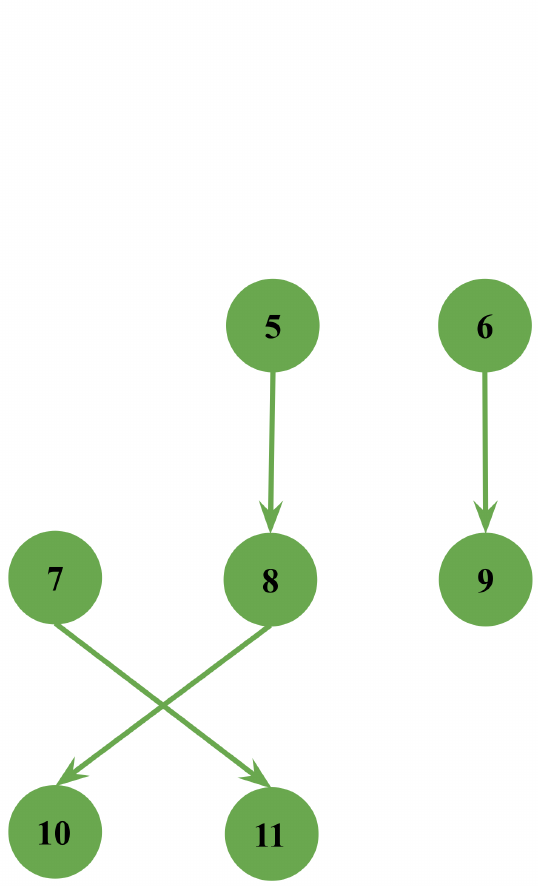} \label{proj_subgraph}
           }
          \hspace{0.5cm}
           \subfloat[Final Graph]{%
              \includegraphics[width=0.2\textwidth]{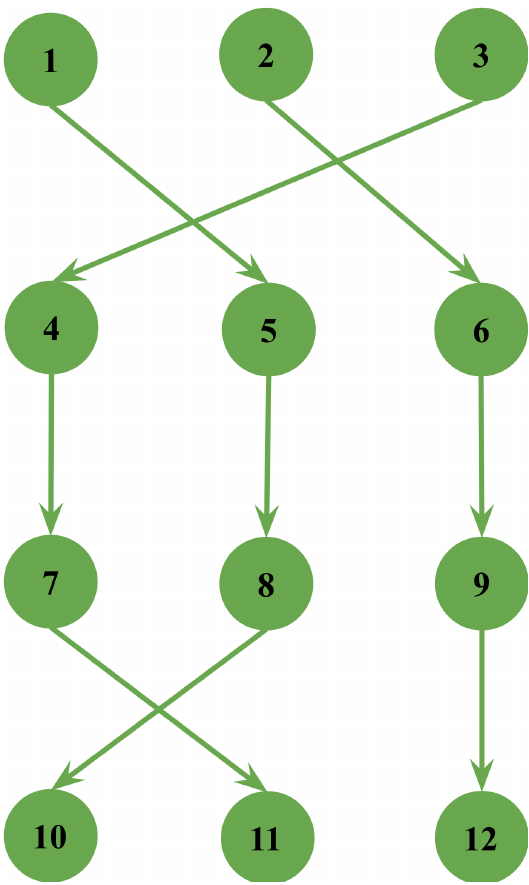} \label{final_round_result}
           }
           \caption{Toy example showing how we round our method's output solutions. \ref{trehsold_result} shows the output of binarizing our method's output by setting a threshold on edge's classification score at 0.5. Edges' arrows indicate time direction. In the example, we observe 4 violated constraints: outgoing flow is greater than 1 for both nodes 5 and 8, and incoming flow is greater than 1 for nodes 9 and 11. \ref{viol_subgraph} shows the subgraph corresponding to the edges involved in these violated constraints, which requires rounding by either our exact or greedy scheme. \ref{proj_subgraph} shows indicates the result of rounding solutions in the subgraph. Hence, it no longer has constraints violations. \ref{final_round_result} shows the final graph, in which both the solutions obtained by thresholding in \ref{trehsold_result}, and the rounding solutions of the subgraph in \ref{final_round_result}  are combined to yield the final trajectories.}\label{rounding_figure}
           
\end{figure*}

As explained in the main paper (Section \ref{inference}), a forward pass through our model yields a fractional solution to the original flow problem with values between 0 and 1.
Thanks to our time-aware message passing network, binarizing this solution directly by setting a threshold at 0.5 will yield a solution that satisfies close to 99\% of the flow conservation constraints on average over test sequences (see Section 5.2 in the main paper).
In order to guarantee that all of them are satisfied, we propose two simple schemes, and describe them in this section. See Figure \ref{rounding_figure} for a summary of our procedure.
\subsection{Greedy Rounding} \label{greedy_rounding_subsec}
\begin{algorithm}[t]  

\DontPrintSemicolon

\SetKwInOut{Input}{Input}
\SetKwInOut{Output}{output}
\caption{Greedy Rounding}
\label{greedy_rounding}

\Input{Graph $G = (V, E)$ \newline
  Fractional solution $\hat{y}$}
\KwResult{Binary feasible solution $y$}
\tcp{Threshold initial solution}
\ForEach{$(i, j) \in E$}{
\If{ $\hat{y}_{(i, j)}\geq 0.5$}{$y_{(i, j)}\leftarrow 1$}
\Else{$y_{(i, j)}\leftarrow 0$}
}
\tcp{Iterate over constraints}
\ForEach{$i \in V$}{
\tcp{Set as inactive all edges but the one with max. score}
\If{Constraint \ref{flow_in_constr} is violated}{
    $j^* \leftarrow \text{argmax}_ {j\in N_i^{past}} \hat{y}_{(i, j)}$ \\

\ForEach{$j\in N_i^{past}\setminus \{j^* \}$}{$y_{(i, j)} \leftarrow 0$}}
\If{Constraint \ref{flow_out_constr} is violated}{
    $j^* \leftarrow \text{argmax}_ {j\in N_i^{fut}} \hat{y}_{(i, j)}$ \\
\ForEach{$j\in N_i^{fut}\setminus \{j^* \}$}{$y_{(i, j)} \leftarrow 0$}}
}
\Return{$y$}
\end{algorithm}

In our setting, having a violated incoming (resp. outgoing) flow conservation constraint means that, for some node, there is more than one  incoming (resp. outgoing) edge classified as active. Hence, a simple way to obtain a binary solution satisfying all constraints is to only set as active the incoming (resp. outgoing) edge with the maximum classification score, for every node.

Let $G=(V, E)$ a MOT graph. Observe that, by following this simple policy, we are guaranteed to obtain a binary feasible solution after $o(\Delta(G)|V|)$ steps, where $\Delta(G)$ indicates the maximum degree of any vertex in $G$. Indeed, observe that we have a total of $2|V|$ constraints, since for each node, there are two flow conservation inequalities. 
Evaluating each of these requires computing a sum of $o(\Delta(G))$ terms, and picking the edge with maximum score among all neighbors in past / future frames has, again, complexity $o(\Delta(G))$. 
Further observe that, by picking the edge with the highest classification score no new constraints can be violated. Indeed, setting all non-maximum incoming (resp. outgoing) edges in a node to zero will make all  remaining left hand sides in inequalities \ref{flow_in_constr} and \ref{flow_out_constr} for other nodes become smaller or equal.  
Hence, it is clear that, at most, $2|V|$ iterations with $o(\Delta(G))$ operations each will be necessary, which yields a total complexity of $o(\Delta(G)|V|)$. See Algorithm \ref{greedy_rounding} for a summary of this procedure.

\subsection{Exact Rounding}
As we show in Table \ref{rounding_tab}, the greedy rounding scheme we just introduced works very well in practice. This is due to the fact that our method's output solutions already satisfy almost all constraints and hence, there is little margin for our rounding scheme to affect performance.
However, in general, greedy rounding is not guaranteed to be optimal. 
 We now explain how \textit{exact} rounding can be performed via linear programming.


Let $\hat{y}\in [0, 1]^{|E|\times 1}$ denote the fractional output of our network at every edge. Also let $A\in  \{0, 1\}^{2|V|\times|E|}$ be the matrix resulting from expressing constraints \ref{flow_in_constr} and \ref{flow_out_constr} for all nodes in matrix notation, and $\mathbbm{1}_{2|V|}$ a $2|V|-$dimensional column vector of ones corresponding to the constraint's right hand side. The exact rounding problem consists in obtaining the binary solution $y_{int}\in \{0, 1\}^{|E|\times 1}$ satisfying constraints \ref{flow_in_constr}   and \ref{flow_out_constr} for all entries that is closest (w.r.t the squared euclidean norm) to our networks' output. Hence, in matrix notation exact rounding can be formulated as:

\begin{align*}
\begin{array}{ll@{}ll}
\min_{y_{int}}  & \lVert y_{int} - \hat{y} \rVert_2^2     &\\
\text{subject to}& Ay_{int} \leq \mathbbm{1}_{2|V|}      &\\                                               &y_{int} \in \{0,1\}^{|E|\times 1}
\end{array}
\end{align*}
However, the quadratic cost can be equivalently written as a linear function. Indeed:
\begin{align*}
\min_{y_{int}} \lVert y_{int} - \hat{y} \rVert_2^2 &= \min_{y_{int}} (y_{int} - \hat{y})^T(y_{int} - \hat{y}) \\
&= \min_{y_{int}} y_{int}^T y_{int} - 2y_{int}^T\hat{y} + \hat{y}^T\hat{y}\\
&= \min_{y_{int}} y_{int}^T y_{int} - 2y_{int}^t\hat{y} \\
&= \min_{y_{int}} y_{int}^T \mathbbm{1}_{|E|} - 2y_{int}^T\hat{y} \\
&= \min_{y_{int}} y_{int}^T (\mathbbm{1}_{|E|} - 2\hat{y})
\end{align*}
Where $\mathbbm{1}_{|E|}$ is an $|E|-$dimensional column vector of ones. Observe that $y_{int}^T y_{int} =  y_{int}^T \mathbbm{1}_{|E|}$ holds due to the fact that $y_{int}$ is a vector of ones and zeros.

Moreover, matrix $A$, is \textit{totally unimodular} \cite{Fleuret:2008:PAMI, networkflows}, which implies that the integrality constraints can be relaxed to box constraints $y_{int} \in [0,1]^{|E|\times 1}$. Therefore, we can relax our original quadratic integer program to a linear program, and solve it in polynomial time while still being guaranteed integer solutions. 

In general, this optimization problem shows a straightforward connection between our setting and the general min-cost flow problem. When naively rounding our solutions with linear programming, we could view our model's output as edge costs in a min-cost-flow problem instance. The key difference is that, in practice, we do not need to solve the entire problem. Since our model's output is almost feasible, when rounding, we can obtain binary solutions for almost all edges by directly thresholding their classification scores. With the remaining edges in which flow conservation constraints are violated, we consider their corresponding subgraph, which typically consists of less than $5\%$ of edges in the graph, and either solve the linear program we have described or use the greedy procedure explained in Subsection \ref{greedy_rounding_subsec}.

\subsection{Performance Comparison} 

In Table \ref{rounding_tab}, we compare the runtime and performance of both rounding schemes with our model's final configuration. Both \textit{exact} and \textit{greedy} rounding show almost equal performance, with greedy rounding having slightly lower IDF1 (0.2 percentage points), equal MOTA and equal speed. Overall this shows, that given the high constraint satisfaction of our model's solutions, the method for rounding has little effect on the tracking results. This is to be expected: since there are few edges in which the rounding procedure needs to be applied, there is little room to affect overall results. Instead, the key element driving performance in our model is our message passing network formulation.

\input{tables/ablation_rounding.tex}


%% file: tables/ablation_rounding.tex

\begin{table} 
\center
\tabcolsep=0.11cm

    \resizebox{\columnwidth}{!}{
    \begin{tabular}{l c c c c c c c c c}
     \toprule
     Rounding & MOTA $\uparrow$ & IDF1 $\uparrow$ & MT $\uparrow$ & ML $\downarrow$ & FP $\downarrow$ & FN $\downarrow$ & ID Sw. $\downarrow$ & Hz. $\downarrow$ \\ [0.5ex] 
     \midrule
     Greedy & 64.0 &  69.8 &  638 &  362 &  5703 &  115089 &  640 & 6.5\\
     Exact & 64.0 & 70.0 & 648 &  362 & 6169 & 114509 & 602 & 6.5\\

     \midrule

    \end{tabular}}

\caption{We compare the effect of having an approximate \textit{greedy} rounding scheme, with respect to performing \textit{exact} rounding via linear programming. Both methods show almost the same performance and no significant difference in speed (Hz, in frames per second).}
\vspace{-0.2cm}
\label{rounding_tab}

\end{table}

%% file: tables/layer_sizes.tex
 \begin{table}[t]
 \centering
 \begin{tabular}{c  c  c  c }
      \toprule

 & \multicolumn{1}{c }{Layer Num.} & \multicolumn{1}{c}{Type} & \multicolumn{1}{c}{Output Size}\\
\midrule
 \parbox[t]{2mm}{\multirow{4}{*}{\rotatebox[origin=c]{90}{\textbf{Feature Encoders} \qquad  \qquad \qquad \qquad }}} 
  & \multicolumn{1}{c}{} & \multicolumn{1}{c}{Nodes  ($\mathcal{N}_v^{enc}$)} & \multicolumn{1}{c}{}\\
 \cline{2-4}
 
  & 0 & \centering \arraybackslash Input  & $3 \times 128 \times 64$ \\
  & 1 & conv $7\times7$ & $64 \times 64 \times 32$ \\
  & 2 & max pool $3 \times 3$ & $64 \times 32 \times 16$ \\
  & 3 & conv1 & $256 \times 32 \times 16$ \\
  & 4 & conv2 & $512 \times 16 \times 8$ \\
  & 5 & conv3 & $1024 \times 8 \times 4$ \\
  & 6 & conv4 & $2048 \times 8 \times 4$ \\
  & 7 & GAP & $2048$ \\
  & 8 & FC + ReLU& $512$ \\
  & 9 & FC + ReLU& $128$ \\
  & 10 & FC + ReLU& $32$ \\
  \cline{2-4}
  & \multicolumn{1}{c}{} & \multicolumn{1}{c}{Edges ($\mathcal{N}_e^{enc}$)} & \multicolumn{1}{c}{}\\
 \cline{2-4}
  & 0 & Input & $6$ \\
  & 1 & FC + ReLU& $18$ \\
  & 2 & FC + ReLU& $18$ \\
  & 3 & FC + ReLU& $16$ \\
 \hline
 
  \hline
 \parbox[t]{2mm}{\multirow{4}{*}{\rotatebox[origin=c]{90}{\textbf{Message Passing Network \quad \quad \quad  \ \ }}}} 

  & \multicolumn{1}{c}{} & \multicolumn{1}{c}{Past Update ($\mathcal{N}^{past}_v$)} & \multicolumn{1}{c}{}\\
 \cline{2-4}
   & 0 & Input & $80$ \\
  & 1 & FC + ReLU& $56$ \\
  & 2 & FC + ReLU & $32$ \\
   \cline{2-4}

    & \multicolumn{1}{c}{} & \multicolumn{1}{c}{Future Update ($\mathcal{N}^{fut}_v$)} & \multicolumn{1}{c}{}\\
 \cline{2-4}
   & 0 & Input & $80$ \\
  & 1 & FC + ReLU& $56$ \\
  & 2 & FC + ReLU & $32$ \\
   \cline{2-4}
  & \multicolumn{1}{c}{} & \multicolumn{1}{c}{Node Update ($\mathcal{N}_v$)} & \multicolumn{1}{c}{}\\
 \cline{2-4}
   & 0 & Input & $64$ \\
  & 1 & FC + ReLU& $32$ \\
  \cline{2-4}
  & \multicolumn{1}{c}{} & \multicolumn{1}{c}{Edge Update ($\mathcal{N}_e$)} & \multicolumn{1}{c}{}\\
 \cline{2-4}
  & 0 & Input & $160$ \\
  & 1 & FC + ReLU & $80$ \\
  & 2 & FC + ReLU & $16$ \\
 \hline


 \hline
 \parbox[t]{2mm}{\multirow{4}{*}{\rotatebox[origin=c]{90}{\textbf{Classifier}  \  }}} 
  & \multicolumn{1}{c}{} & \multicolumn{1}{c}{Edges ($\mathcal{N}_e^{class}$)} & \multicolumn{1}{c}{}\\
 \cline{2-4}
   & 0 & Input & $16$ \\
  & 1 & FC + ReLU & $8$ \\
  & 2 & FC + Sigmoid & $1$ \\
\bottomrule
 \end{tabular}

  \caption{For each network component, we specify its number of layers and input / output dimensions. FC denotes a \textit{fully connected} layer and GAP, \textit{Global Average Pooling}. . For the Node Encoder, $\mathcal{N}_v^{enc}$, all layers up to position 7 correspond to those of a ResNet50\cite{He2016DeepRL}. Hence, `layers' 3 to 6 are actually sequences of residual blocks. The only modification we have applied is using stride 1 in all convolutional layers of conv4. Hence, there is no spatial size reduction from conv3 to conv4. } \label{tab:layer_sizes}
 \end{table}
 

%% file: sections/further_implementation_details.tex
\section{Further Implementation Details} \label{futher_impl_details_section}
In this section, we extend the information provided in the main article about the implementation of our method.

\subsection{Detailed Architecture}
 In Table \ref{tab:layer_sizes}, we specify the configuration of each of the network's components. Observe that our model is composed of a total of 6 networks. The first two, $\mathcal{N}_v^{enc}$ and $\mathcal{N}_e^{enc}$,  are used for feature encoding  of nodes and edges, respectively (see section 4.3 in the main paper). For neural message passing, we use one network to update edge embeddings,  $\mathcal{N}_e$, and three networks to update node embeddings $\mathcal{N}_v^{past}$, $\mathcal{N}_v^{fut}$ and $\mathcal{N}_v$ (see sections 4.1  and 4.2 in the main paper). Lastly, to classify edges, we use another network, $\mathcal{N}_e^{class}$ (see section 4.4 in the main paper).

\subsection{Batch Processing}

As explained in the main paper, we process videos by sequentially feeding overlapping batches of 15 frames to our model. In the MOTChallenge, different sequences show great variability regarding (i) number of frames per second at which videos are recorded (ii) presence of camera movement and (iii) number of detections per frame. To account for (i) and (i), we sample a fixed and number of frames per second for static and dynamic sequences, which we set to 6 and 9, respectively. To tackle (iii), we restrict the connectivity of graphs by connecting two nodes only if both are among the top-50 reciprocal nearest neighbors according to the ResNet features. This ensures that our model scales to crowded sequences with no significant overhead, and that the topology of our graphs is comparable among different videos.

\subsection{Cross-Validation Splits} \label{cross_val_splits}
We conduct all of our experiments with 3-fold cross-validation on the MOT17 benchmark training data. To do so, we split the sequences into three subsets.  
For each experiment configuration we train a total of 3 networks: one for each possible validation set. Since our splits cover all training sequences of MOT17, we obtain metrics over the whole dataset which allow us to choose the best network configuration and set of hyperparameters.

In Table \ref{tab:split},  we report the validation sequences corresponding to each split. For each of the splits, the sequences not contained in its validation set are used for training, together with those of the MOT15 dataset. 

  When deciding which sequences to include in each split, we made sure that each subset contains both moving and static camera sequences. Furthermore, we balance the number of tracks and sequence length in seconds (recall that fps is normalized during processing) in each split, in order to ensure that all validation settings are comparable.

\input{tables/splits.tex} 

\subsection{Preprocessing and Postprocessing} \label{preprocessing}

As we explain in the next section we use \cite{tracktor} to preprocess public detections. As an alternative, for the results in Section \ref{graph_meth_comparison_section}, we follow a similar detection preprocessing scheme to the one applied by other methods \cite{Chen2018RealTimeMP, Yoon2018OnlineMT, tracking_corr_clust}. 
We use both the bounding box regressor and classifier heads of a Faster-RCNN\cite{faster_rcnn} trained on the MOT17 Detection challenge. 
We filter out all bounding boxes whose confidence score is smaller than 0.5, and correct the remaining with the bounding box regressor. 
After that, we apply standard Non-Maxima-Supression to the resulting boxes, by using each box' confidence score, and setting an IoU threshold of 0.85.
For post-processing, if using \cite{tracktor} we fill gaps in our trajectories by matching our output trajectories to the ones in \cite{tracktor}, and then using the latter to fill the detections in missing frames. For the remaining missing gaps in our trajectories, we use bilinear interpolation. Finally, we drop all trajectories that consist of a single detection. This allows our model to identify false positives as \textit{isolated} nodes in the graph (i.e. nodes with neither incoming nor outgoing active edges).

\subsection{Baseline} \label{baseline}
As explained in the main paper, we use \cite{tracktor} as a baseline. More specifically, we preprocess all sequences by first running \cite{tracktor} on public detections. After that, we discard the pedestrian ID assigned by \cite{tracktor}, and simply treat the resulting boxes as raw detections for our neural solver. \cite{tracktor} uses the regression head of a Faster-R-CNN \cite{faster_rcnn} in order to predict the next locations of objects in neighboring frames 

Other graph approaches resort to low-level image features and work with raw (i.e. non-maxima suppressed) detections to approach this challenge \cite{henscheltpami2016, tangbmtt, tang2015subgraph, People_Tracking, tracking_corr_clust}. Observe that using raw detections has indeed, more potential than just adding neighboring detections with \cite{tracktor}, as it yields a greatly increased number of object hypothesis. Hence, it allows tracking methods to have the capacity to track more objects. However, \cite{tracktor} reduces computational times significantly, and provides more precise boxes, which improves the efficiency of our method.We perform a detailed comparison with graph-based methods in the next section.

%% file: tables/splits.tex
\begin{table}[h]
\center
\tabcolsep=0.11cm

    \resizebox{\columnwidth}{!}{
    \begin{tabular}{l c c c c c}
     \toprule
     Name & Mov. & Length (s) & Length (f) & Tracks & Boxes \\ [0.5ex] 
     \midrule
     \multicolumn{6}{c}{Split 1} \\
     \midrule
      MOT17-02 &        No &      20   &     600      &   62  & 18581 \\
      MOT17-10 &       Yes &      22   &     654     &   57  & 12839 \\
      MOT17-13 &       Yes &      30   &     750     &   110  & 11642 \\
      \hline
      Overall &  -- &      72   &     2004     &   229  & 43062 \\
     \midrule
     \multicolumn{6}{c}{Split 2} \\
     \midrule
      MOT17-04 &       No   &     35    &     1050    &   83  & 47557 \\
      MOT17-11 &      Yes  &     30    &      900    &   75  & 9436 \\
      \hline
      Overall &  -- &      65   &     1950     &   158  & 56993 \\
      \midrule
     \multicolumn{6}{c}{Split 3 } \\
     \midrule
      MOT17-05 &       Yes &      60   &     837     &   133  & 6917 \\
      MOT17-09 &       No  &      18   &     525     &   26  & 5325 \\
      \hline
      Overall &  -- &      78   &     1362     &   159  & 12242 \\
     \bottomrule
     \multicolumn{6}{c}{Total } \\
     \midrule
        &       -- &      205   &   5316  &   546  &  112297 \\
     \bottomrule     
     
    \end{tabular}}

\caption{We report the sequences used in each cross-validation split in order to evaluate our model. \textit{Mov} refers to whether there is camera movement in the scene and, for length 's' denotes seconds and 'f', frames. Note that in the MOT17 benchmark, each sequence is given with three sets of detections (DPM, FRCNN and SDP). Since the ground truth does not change among them, here we report the features of each sequence, and do not take into account the set of detections. However, when testing our models, we make use of all sets of  detections.}
\vspace{-0.2cm}
\label{tab:split}

\end{table}

%% file: sections/no_tracktor_metrics.tex
\section{Additional Comparison with Graph Methods} \label{graph_meth_comparison_section}
\input{tables/mot_graph_meth.tex} 

We provide an extended comparison of our method\footnote{We made a slight change in the configuration of our method for these results. Since we do not have access to \cite{tracktor} and, hence, we have to rely heavily on linear interpolation for postprocessing (see Appendix \ref{preprocessing}), we augment the frame sampling rate at which we process sequences, and also the size of graphs we process proportionally, in order to cover time intervals of the same size of those of our main configuration. Specifically, we increase the sampling rate of frames for static sequences from 6 to 9, and from 9 to 15 for those with a moving camera. As for the number of frames corresponding to each processed graph, we increase it from 15 to 25.} with top-performing offline graph-based methods. The results are summarized in Table \ref{tab:mot_graph_meth}. For each method, we highlight the additional features and sources of information that it has access to. Additionally, we provide the results obtained by our method when we do not use our baseline \cite{tracktor} for preprocessing detections, and we denote it with \textit{Ours*}. We show that, even in that case, our method still surpasses previous works by a significant margin even though it has access to significantly less information. Hence, these results  further confirm the superiority of our approach. 

Even without \cite{tracktor}, in the MOT15\cite{MOTChallenge:arxiv:2015} dataset we observe an improvement of 19.8 points in MOTA and 15.6 points in IDF1 with respect to \cite{icra_net_flows}, which uses the same underlying Min-Cost Flow graph formulation, but a simpler learning scheme. Moreover, in all three datasets, our method consistently improves significantly upon multi-cut based methods \cite{jCCpami2018, maACCV2019,tracking_corr_clust, TangAAS17}, which use a more involved graph formulation, have access to a significantly larger number of boxes due to not using Non-Maximum Suppression, and either employ low-level image features or use a hierarchical scheme. Thus, we clearly demonstrate that our method shows very strong performance and surpasses previous work, even when it cannot leverage low-level image information via \cite{tracktor}. Furthermore, when our method is given access to additional features as other methods, it shows its full potential and outperforms all previous works by an even larger margin.



%% file: tables/mot_graph_meth.tex

\begin{table*}
\center
\tabcolsep=0.11cm

    \begin{tabular}{l c c c c c c c c c }
     \toprule
     Method & MOTA $\uparrow$ & IDF1 $\uparrow$ & MT $\uparrow$ & ML $\downarrow$ & ID Sw. $\downarrow$  & Hz  $\uparrow$ & Additional Features \\ [0.5ex] 
     \midrule
     \multicolumn{8}{c}{2D MOT 2015~\cite{lealarxiv2015}    } \\
     \midrule
        Ours & \textbf{51.5} &  \textbf{58.6} & \textbf{31.2} &  \textbf{25.9}  & \textbf{375} & 6.5 & Box regression \cite{tracktor} \\
          Ours* & 46.6 &  53.8 & 25.2 &  29.0  & 381 & 6.5 & --------- \\

     JointMC~\cite{jCCpami2018} & 35.6 & 45.1 & 23.2 & 39.3 & 457 & 0.6 & Point trajectories \cite{point_trajectories}, opt. flow\\

    QuadMOT~\cite{Son_2017_CVPR} & 33.8 & 40.4 & 12.9 & 36.9 & 703 & 3.7 & Learnable box regression\\

    MHT\_DAM~\cite{kimiccv2015} & 32.4 & 45.3 & 16.0 & 43.8 & 435 & 0.7 &  ---------\\

    MCF\_PHD~\cite{icra_net_flows} & 29.9 & 38.2 & 11.9 & 44.0 & 656 & 12.2 &  --------- \\

    DeepFlow \cite{Schulter_2017_CVPR}& 26.8 & -- & -- & -- & -- & -- & ALFD motion features \cite{Choi2015}\\

     \midrule
     \multicolumn{8}{c}{MOT16~\cite{milanarxiv2016}} \\
     \midrule
     Ours &  \textbf{58.6} & \textbf{61.7} & \textbf{27.3} &  \textbf{34.0} & \textbf{354} & 6.5  & Box regression \cite{tracktor} \\
    
     Ours* &  54.3 & 56.8 & 23.5 &  36.0 &  440 & 6.5 &  ---------\\

     NOTA~\cite{aggregate_Track_app}  & 49.8 &	55.3 &	17.9 & 37.7 & 614 & -- & Not public\\
     
     HCC~\cite{maACCV2019} & 49.3 & 50.7 & 17.8 & 39.9 & 391  & 0.8 & Tracklets,  Deep Matching \cite{deep_matching}\\

     LMP~\cite{TangAAS17} & 48.8 & 51.3 & 18.2 & 40.1 & 481  & 0.5 & Body part detections, Deep Matching \cite{deep_matching}, no-NMS\\

     TLMHT~\cite{soa_mht} & 48.7 & 55.3 & 15.7 &  44.5  & 413 & 4.8 & Tracklets \\
     FWT~\cite{HenschelLCR17} & 47.8 & 44.3 & 19.1 & 38.2 & 852 & 0.6 & Head detections, Deep Matching \cite{deep_matching} \\

     \midrule
     \multicolumn{8}{c}{ MOT17~\cite{milanarxiv2016}   } \\
     \midrule  
     Ours & \textbf{58.8} & \textbf{61.7} & \textbf{28.8} & \textbf{33.5} & \textbf{1185} & 6.5 & Box regression \cite{tracktor}\\
     
     Ours* & 56.0 & 58.4 & 26.2 & 34.7 &  1451 & 6.5 & ---------\\

     JBNOT \cite{JBNOT} & 52.6 & 50.8 & 19.7 & 35.8 & 3050 & 5.4 & Joint detections, Deep Matching \cite{deep_matching}\\

     eHAF\cite{sheng2018}     & 51.8 & 54.7 & 23.4 & 37.9 & 1834 & 0.7 & Superpixels, Segmentation, Foreground Extraction \\
     NOTA \cite{aggregate_Track_app}  & 51.3 &	54.7 &	17.1 & 35.4 & 2285 & -- & Not public\\
     FWT~\cite{HenschelLCR17} & 51.3 & 47.6 & 21.4 & 35.2 & 2648 & 0.2 &  Head detections, Deep Matching \cite{deep_matching}\\
     jCC~\cite{jCCpami2018} & 51.2 & 54.5   & 20.9 & 37.0 &  1802 & 1.8 & Point trajectories \cite{point_trajectories}, opt. flow\\


     \bottomrule
    \end{tabular}

\caption{We compare both our final method (Ours) and a variant of our method in which we do not exploit our baseline \cite{tracktor} (Ours*) to other top-performing graph-based offline methods in the MOTChallenge benchmark. Under \textit{Additional Features}, we highlight which information sources or features each method used apart from the given public detections. We denote with a horizontal line the absence of such features. For \cite{aggregate_Track_app}, we cannot provide this information, since the article is not freely available. For \cite{Schulter_2017_CVPR}, we report the only metric that the authors reported in their article. We still include it the table due to the fact that it also follows the min-cost flow MOT formulation, and is similar in spirit to our work (see Related Work in the main article). }
\vspace{-0.2cm}
\label{tab:mot_graph_meth}

\end{table*}